\documentclass[10pt,twocolumn,letterpaper]{article}

\usepackage{cvpr}
\usepackage{times}
\usepackage{epsfig}
\usepackage{graphicx}
\usepackage{amsmath}
\usepackage{amssymb}
\usepackage{multirow}
\usepackage{subfigure}
\usepackage{pifont}
\usepackage{xcolor}
\usepackage{array}
\usepackage{url}

\newcommand{\xmark}{\ding{55}}
\newcommand{\rmark}{\ding{52}}

\newcolumntype{I}{!{\vrule width 1.2pt}}
\newlength\savedwidth
\newcommand\whline{\noalign{\global\savedwidth\arrayrulewidth
		\global\arrayrulewidth 1.25pt}%
	\hline
	\noalign{\global\arrayrulewidth\savedwidth}}

\cvprfinalcopy 

\def\cvprPaperID{3580} 

\ifcvprfinal\fi
\begin{document}
	
	\title{Learning from Synthetic Data for Crowd Counting in the Wild}
	
	\author{Qi Wang,\qquad Junyu Gao,\qquad Wei Lin,\qquad Yuan Yuan\\
		School of Computer Science and Center for OPTical IMagery Analysis and Learning (OPTIMAL),\\
		Northwestern Polytechnical University, Xi'an, Shaanxi, P. R. China\\
		{\tt\small \{crabwq, gjy3035, elonlin24, y.yuan1.ieee\}@gmail.com}
	}
	
	\maketitle

	\begin{abstract}
		Recently, counting the number of people for crowd scenes is a hot topic because of its widespread applications (e.g. video surveillance, public security). It is a difficult task in the wild: changeable environment, large-range number of people cause the current methods can not work well. In addition, due to the scarce data, many methods suffer from over-fitting to a different extent. To remedy the above two problems, firstly, we develop a data collector and labeler, which can generate the synthetic crowd scenes and simultaneously annotate them without any manpower. Based on it, we build a large-scale, diverse synthetic dataset. Secondly, we propose two schemes that exploit the synthetic data to boost the performance of crowd counting in the wild: 1) pretrain a crowd counter on the synthetic data, then finetune it using the real data, which significantly prompts the model's performance on real data; 2) propose a crowd counting method via domain adaptation, which can free humans from heavy data annotations. Extensive experiments show that the first method achieves the state-of-the-art performance on four real datasets, and the second outperforms our baselines. The dataset and source code are available at \url{https://gjy3035.github.io/GCC-CL/}.
		
	\end{abstract}
	
	\section{Introduction}
	\label{intro}
	
	Crowd counting is a branch of crowd analysis \cite{DBLP:journals/spm/JuniorMJ10,DBLP:conf/iccv/RodriguezSLA11,DBLP:conf/aaai/LiCNW17,wang2018detecting}, which is essential to video surveillance, public areas planning, 
	traffic flow monitoring and so on. This task aims to predict density maps and estimate the number of people for crowd scenes. At present, many CNN- and GAN-based methods \cite{zhang2016single,sam2017switching,shen2018crowd,shi2018crowd,cao2018scale} attain a phenomenal performance on the existing datasets. The above methods focus on how to learn effective and discriminative features (such as local patterns, global contexts, multi-scale features and so on) to improve models' performance.
	
	\begin{figure}
		\centering
		\includegraphics[width=0.45\textwidth]{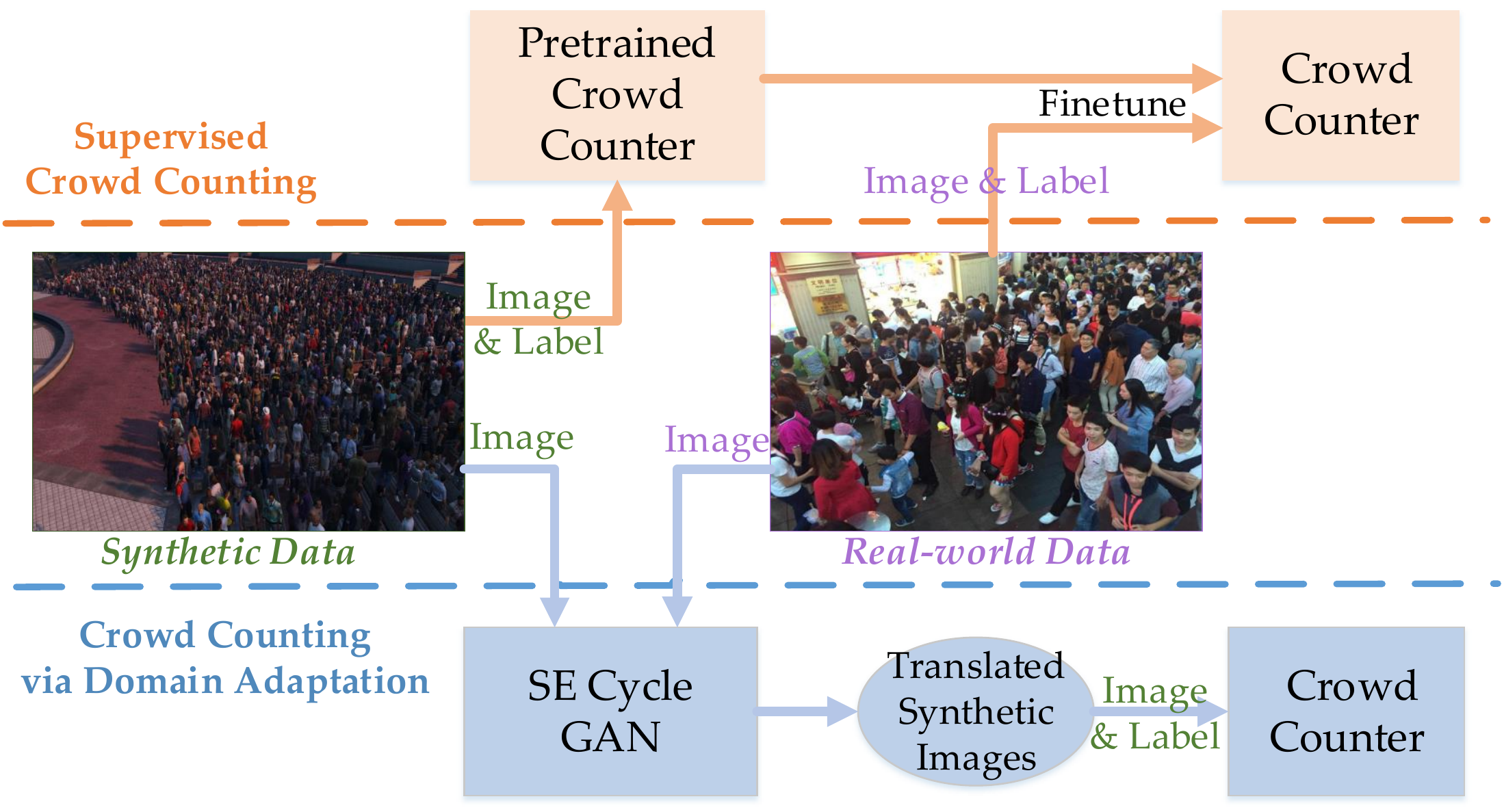}
		\caption{Two ways of using the proposed GCC dataset: supervised learning and domain adaptation. }\label{Fig-intro}
		\vspace{-0.45cm}
	\end{figure}
	
	At the same time, The aforementioned mainstream deep learning methods need a large amount of accurately labeled and diversified data. Unfortunately, current datasets \cite{chan2008privacy, chen2012feature, zhang2016data, zhang2016single,Qi2017Deep,idrees2013multi,idrees2018composition} can not perfectly satisfy the needs, which also results in two intractable problems for crowd counting in the wild. Firstly, it causes that the existing methods cannot be performed to tackle some unseen extreme cases in the wild (such as changeable weather, variant illumination and a large-range number of people). Secondly, due to rare labeled data, many algorithms suffer from overfitting, which leads to a large performance degradation during transferring them to the wild or other scenes. In addition, there is an inherent problem in the congested crowd datasets: the labels are not very accurate, such as some samples in UCF\_CC\_50 \cite{idrees2013multi} and Shanghai Tech A \cite{zhang2016single} (``SHT A'' for short). 
	
	In order to remedy the aforementioned problems, we start from two aspects, namely data and methodology. From the data perspective, we develop a data collector and labeler, which can generate synthetic crowd scenes and automatically annotate them. By the collector and labeler, we construct a large-scale and diverse synthetic crowd counting dataset. The data is collected from an electronic game Grand Theft Auto V (GTA5), thus it is named as ``GTA5 Crowd Counting'' (``GCC'' for short) dataset. Compared with the existing real datasets, there are four advantages: 1) free collection and annotation; 2) larger data volume and higher resolution; 3) more diversified scenes and 4) more accurate annotations. The detailed statistics are reported in Table \ref{Table-compare}.

	From the methodological perspective, we propose two ways to exploit synthetic data to improve the performance in the wild. Firstly, we propose a supervised strategy to reduce the overfitting phenomenon. To be specific, we firstly exploit the large-scale synthetic data to pretrain a crowd counter, which is our designed Spatial Fully Convolutional Network (SFCN). Then we finetune the obtained counter using the real data. This strategy can effectively prompt the performance on real data. Traditional models (training from scratch \cite{zhang2016single,ranjan2018iterative,cao2018scale} or image classification model \cite{babu2018divide,shi2018crowd,idrees2018composition}) have some layers with random initialization or a regular distribution, which is not a good scheme. Compared with them, our strategy can provide more complete and better initialization parameters.

	Secondly, we propose a domain adaptation crowd counting method, which can improve the cross-domain transfer ability. To be specific, we present an SSIM Embedding (SE) Cycle GAN, which can effectively translate the synthetic crowd scenes to real scenes. During the training process, we introduce the Structural Similarity Index (SSIM) loss. It is a penalty between the original image and reconstructed image through the two generators. Compared with the original Cycle GAN, the proposed SE effectively maintains local patterns and texture information, especially in the extremely congested crowd region and some backgrounds. Finally, we translate the synthetic data to photo-realistic images. Based on these data, we train a crowd counter without the labels of real data, which can work well in the wild. Fig. \ref{Fig-intro} demonstrates two flowcharts of the proposed methods.

	In summary, this paper's contributions are three-fold:
	\begin{enumerate}
		\vspace{-0.15cm}
		\item[1)] We are the first to develop a data collector and labeler for crowd counting, which can automatically collect and annotate images without any labor costs. By using them, we create the first large-scale, synthetic and diverse crowd counting dataset. 
		\vspace{-0.15cm}
		\item[2)] We present a pretrained scheme to facilitate the original method's performance on the real data, which can more effectively reduce the estimation errors compared with random initialization and ImageNet model. Further, through the strategy, our proposed SFCN achieves the state-of-the-art results.
		\vspace{-0.15cm}
		\item[3)] We are the first to propose a crowd counting method via domain adaptation, which does not use any label of the real data. By our designed SE Cycle GAN, the domain gap between the synthetic and real data can be significantly reduced. Finally, the proposed method outperforms the two baselines. 
	\end{enumerate}

	\section{Related Works}
	
	\begin{table}[htbp]	
		\centering
		\caption{Statistics of the seven real-world datasets and the synthetic GCC dataset.}
		\scriptsize
		\setlength{\tabcolsep}{0.8mm}{
			\begin{tabular}{c|c|c|c|c|c|c}
				\whline
				\multirow{2}{*}{Dataset}	&Number &Average &\multicolumn{4}{|c}{Count Statistics}\\
				\cline{4-7} 
				& of Images &Resolution & Total &Min & Ave & Max	\\
				\whline
				UCSD  \cite{chan2008privacy}  &2,000 &$158 \times 238$  & 49,885 &11 & 25 & 46	\\
				\hline
				Mall  \cite{chen2012feature}  &2,000 &$480 \times 640$  & 62,325 &13 & 31 & 53	\\
				\hline
				UCF\_CC\_50 \cite{idrees2013multi}  &50 &$2101 \times 2888$  & 63,974 &94 & 1,279 & 4,543	\\
				\hline
				WorldExpo'10\cite{zhang2016data} &3,980 &$576 \times 720$  & 199,923 &1 & 50 & 253 \\
				\hline
				SHT A \cite{zhang2016single}  &482 &$589 \times 868 $  & 241,677 &33 & 501 & 3,139\\
				\hline
				SHT B \cite{zhang2016single}  &716 &$768 \times 1024$  & 88,488 &9 & 123 & 578 \\
				\hline
				UCF-QNRF \cite{idrees2018composition} &1,525 &$2013 \times 2902$  & 1,251,642 & 49 & 815 & 12,865 \\
				\whline
				\textbf{GCC}  &\textbf{15,212} &$\boldsymbol{1080 \times 1920}$  &\textbf{7,625,843}  &\textbf{0} &\textbf{501} &\textbf{3,995}  \\
				\whline
				
			\end{tabular}\label{Table-compare}
		}
		\vspace{-0.5cm}
	\end{table}
	
	\textbf{Crowd Counting Methods.} Mainstream CNN-based crowd counting methods \cite{zhang2015cross,zhang2016single,sindagi2017cnn,sindagi2017generating,li2018csrnet,liu2018leveraging,idrees2018composition,cao2018scale,shi2018crowd,ranjan2018iterative} yield the new record by designing the effective network architectures. \cite{zhang2015cross,sindagi2017cnn} exploit multi-task learning to explore the relation of different tasks to improve the counting performance. \cite{zhang2016single,idrees2018composition,cao2018scale,ranjan2018iterative} integrate the features of multi-stream, multi-scale or multi-stage networks to improve the quality of density maps.  \cite{sindagi2017generating,li2018csrnet} attempt to encode the large-range contextual information for crowd scenes. In order to tackle scarce data, \cite{liu2018leveraging} proposes a self-supervised learning to exploit unlabeled web data, and \cite{shi2018crowd} presents a deep negative correlation learning to reduce the over-fitting.

	\textbf{Crowd Counting Datasets.} In addition to the algorithms, the datasets potentially promote the development of crowd counting. UCSD \cite{chan2008privacy} is the first crowd counting dataset released by Chan \emph{et al.} from University of California San Diego. It records the crowd in a pedestrian walkway, which is a sparse crowd scene. Chen \emph{et al.} \cite{chen2012feature} propose a public Mall dataset which records a shopping mall scene. Idrees \emph{et al.} \cite{idrees2013multi} release the UCF\_CC\_50 dataset for highly congested crowd scenes. WorldExpo'10 dataset is proposed by Zhang \emph{et al.} in \cite{zhang2016data}, which is captured from surveillance cameras in Shanghai 2010 WorldExpo. Zhang \emph{et al.} \cite{zhang2016single} present ShanghaiTech Dataset, including the high-quality real-world images. Idrees \emph{et al.} \cite{idrees2018composition} propose a large-scale extremely congested dataset. More detailed information about them is listed in Table \ref{Table-compare}.
	
	
	\begin{figure*}
		\centering
		\includegraphics[width=1\textwidth]{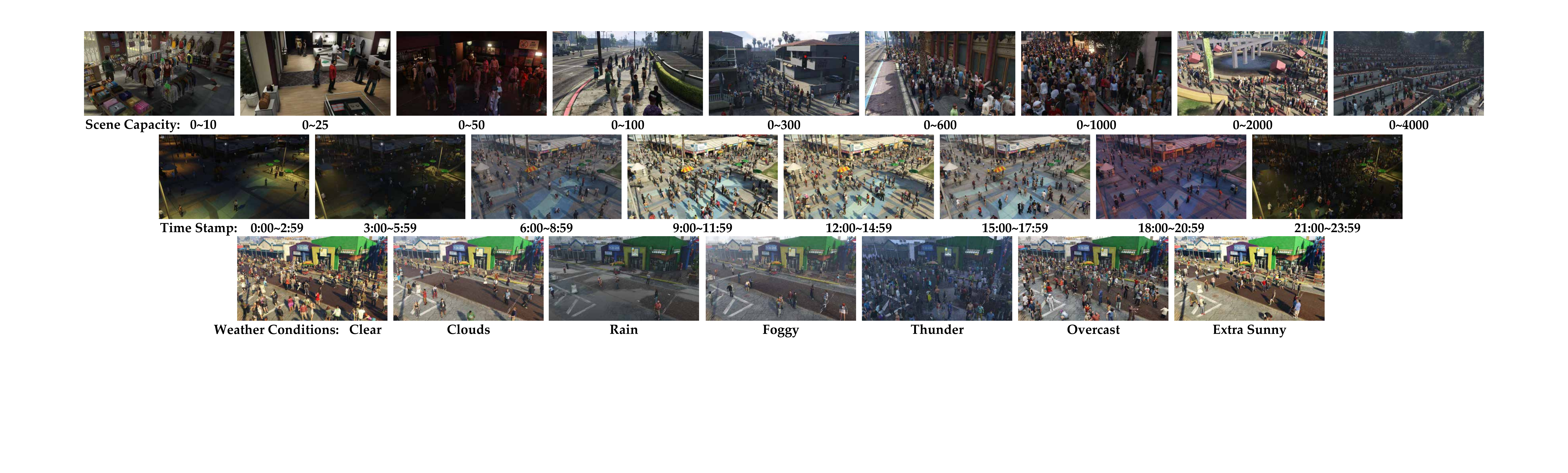}
		\caption{The display of the proposed GCC dataset from three views: scene capacity, timestamp and weather conditions. }\label{Fig-propery}
		\vspace{-0.25cm}
	\end{figure*}
	
	\textbf{Synthetic Dataset.}
	Annotating the groundtruth is a time-consuming and labor-intensive work, especially for pixel-wise tasks (such as semantic segmentation, density map estimation). To remedy this problem, some synthetic datasets \cite{Richter_2016_ECCV,Johnson-Roberson:2017aa,richter2017playing,ros2016synthia,bak2018domain} are released to save the manpower. \cite{Richter_2016_ECCV,Johnson-Roberson:2017aa,richter2017playing} collect synthetic scenes based on GTA5. To be specific, \cite{Richter_2016_ECCV} develops a fast annotation method based on the rendering pipeline. Johnson-Roberson \emph{et al.} \cite{Johnson-Roberson:2017aa} present a method to analyze the internal engine buffers according the depth information, which can produce the accurate object masks. \cite{richter2017playing} proposes an approach to extract data without modifying the source code and content from GTA5, which can provide six types groundtruth. \cite{ros2016synthia,bak2018domain} build synthetic models based on some open-source game engine. \cite{ros2016synthia} exploits Unity Engine \cite{unity} to construct the synthetic street scenes data for autonomous driving, which generates the pixel-wise segmentation labels and depth maps. \cite{bak2018domain} develops a synthetic person re-identification dataset based on Unreal Engine 4 \cite{unreal}.
	

	\section{GTA5 Crowd Counting (GCC) Dataset}

	\label{gcc-dataset}
	
	Grand Theft Auto V (GTA5) is a computer game published by Rockstar Games \cite{rstar} in 2013. In GTA5, the players can immerse themselves into the game in a virtual world, the fictional city of Los Santos, based on Los Angeles. GTA5 adopts the proprietary Rockstar Advanced Game Engine (RAGE) to improve its draw distance rendering capabilities. Benefiting from the excellent game engine, its scene rendering, texture details, weather effects and so on are very close to the real-world conditions. In addition, Rockstar Games allows the players to develop the mod for noncommercial or personal use. 
	
	Considering the aforementioned advantages, we develop a data collector and labeler for crowd counting in GTA5, which is based on Script Hook V \cite{hookv}. Script Hook V is a C++ library for developing game plugins. Our data collector constructs the complex and congested crowd scenes via exploiting the objects of virtual world. Then, the collector captures the stable images from the constructed scenes. Finally, by analyzing the data from rendering stencil, the labeler automatically annotates the accurate head locations of persons.
	
	Previous synthetic GTA5 datasets \cite{Richter_2016_ECCV,Johnson-Roberson:2017aa,richter2017playing} capture normal scenes directed by the game programming. Unfortunately, there is no congested scene in GTA5. Thus, we need to design a strategy to construct crowd scenes, which is the most obvious difference with them. 
	
	\subsection{Data Collection}
	
	\begin{figure}
		\centering
		\includegraphics[width=0.45\textwidth]{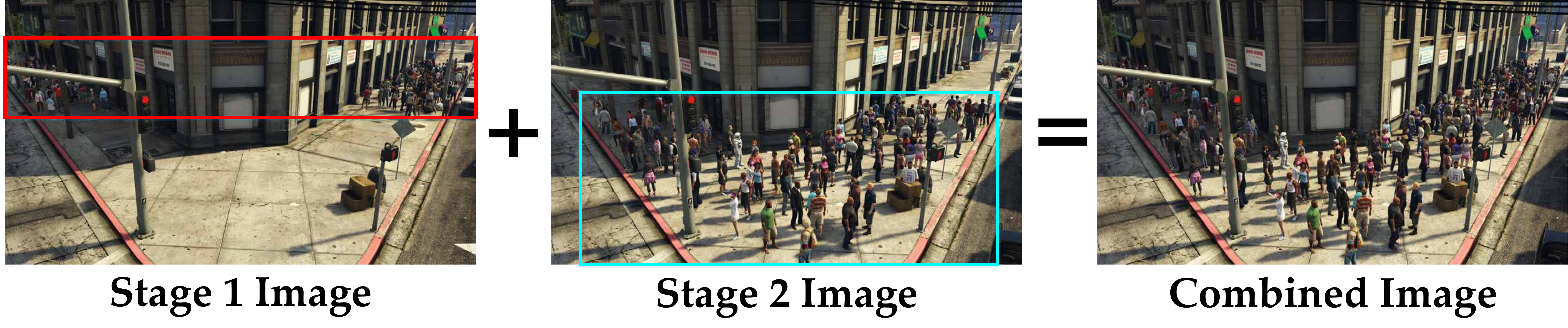}
		\caption{The demonstration of image combination for congested crowd scenes. }\label{Fig-gen}
		\vspace{-0.35cm}
	\end{figure}
	
	This section describes the pipeline of data collection, which consists of three modules as follows.
	
	\textbf{Scene Selection.} The virtual world in GTA5 is built on a fictional city, which covers an area of 252 square kilometers. In the city, we selected 100 typical locations, such as beach, stadium, mall, store and so on. For each location, the four surveillance cameras are equipped with different parameters (location, height, rotation/pitch angle). Finally, the 400 diverse scenes are built. In these scenes, we elaborately define the Region of Interest (ROI) for placing the persons and exclude some invalid regions according to common sense.
	
	\textbf{Person Model.}
	Persons are the core of crowd scenes. Thus, it is necessary that we describe the person model in our proposed dataset. In GCC dataset, we adopt the 265 person models in GTA5: different person model has different skin color, gender, shape and so on. Besides, for each person model, it has six variations on external appearance, such as clothing, haircut, etc. In order to improve the diversity of person models, each model is ordered to do a random action in the sparse crowd scenes. 
	
	\textbf{Scenes Synthesis for Congested Crowd.} Due to the limitation of GTA5, the number of people must be less than 256. Considering this, for the congested crowd scenes, we adopt a step-by-step method to generate scenes. To be specific, we segment several non-overlapping regions and then place persons in each region. Next, we integrate multiple scenes into one scene. Fig. \ref{Fig-gen} describes the main integration process: the persons are placed in the red and green regions in turn. Finally, the two images are combined in the one. 
	
	\textbf{Summary.} The flowchart of generation is described as follows. \emph{Construct scenes}: a) select a location and setup the cameras, b) segment Region of interest (ROI) for crowd, c) set weather and time. \emph{Place persons}: a) create persons in the ROI and get the head positions, b) obtain the person mask from stencil, c) integrate multiple images into one image, d) remove the positions of occluded heads. The demonstration video is available at: \url{https://www.youtube.com/watch?v=Hvl7xWkIueo}.

	\subsection{Properties of GCC}
	
	GCC dataset consists of 15,212 images, with resolution of $1080 \times 1920$, containing 7,625,843 persons. Compared with the existing datasets, GCC is a more large-scale crowd counting dataset in both the number of images and the number of persons. Table \ref{Table-compare} compares the basic information of GCC and the existing datasets. In addition to the above advantages, GCC is more diverse than other real-world datasets.

\begin{figure}
		\centering
		\includegraphics[width=0.48\textwidth]{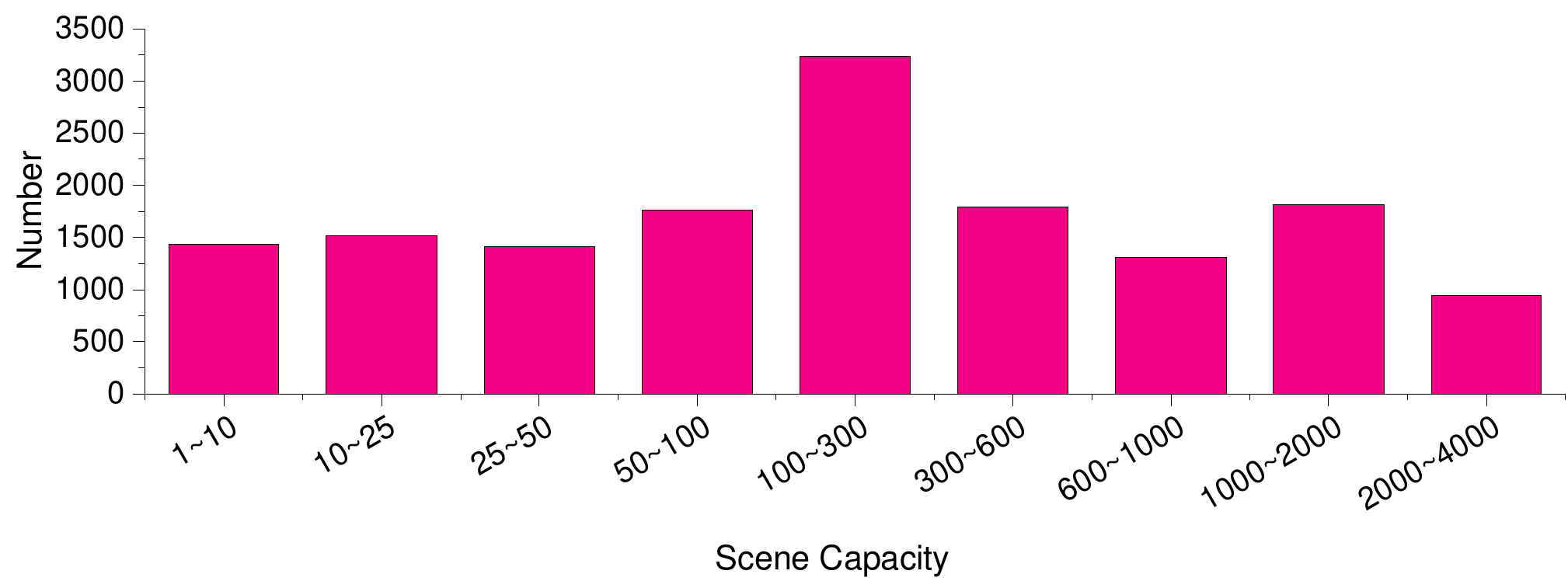}
		\caption{The statistical histogram of crowd counts on the proposed GCC dataset. }\label{Fig-hist}
		\vspace{-0.25cm}
	\end{figure}

	\textbf{Diverse Scenes.} GCC dataset consists of 400 different scenes, which includes multiple types of locations. For example, indoor scenes: convenience store, pub, etc. outdoor scenes: mall, street, plaza, stadium and so on. Further, all scenes are assigned with a level label according to their space capacity.  The first row in Fig. \ref{Fig-propery} shows the typical scenes with different levels. In general, for covering the range of people, the larger scene has more images. Thus, the setting is conducted as follows: the scenes with the first/second/last three levels contain $30/40/50$ images. Besides, the images that contain some improper events should be deleted. Finally, the number of images in some scenes may be less than their expected value. Fig. \ref{Fig-hist} demonstrates the population distribution histogram of our GCC dataset. 
	
	\begin{figure} 
		\centering 
		\subfigure[Time stamp distribution.] { \label{Fig-sector-time} 
			\includegraphics[width=0.46\columnwidth]{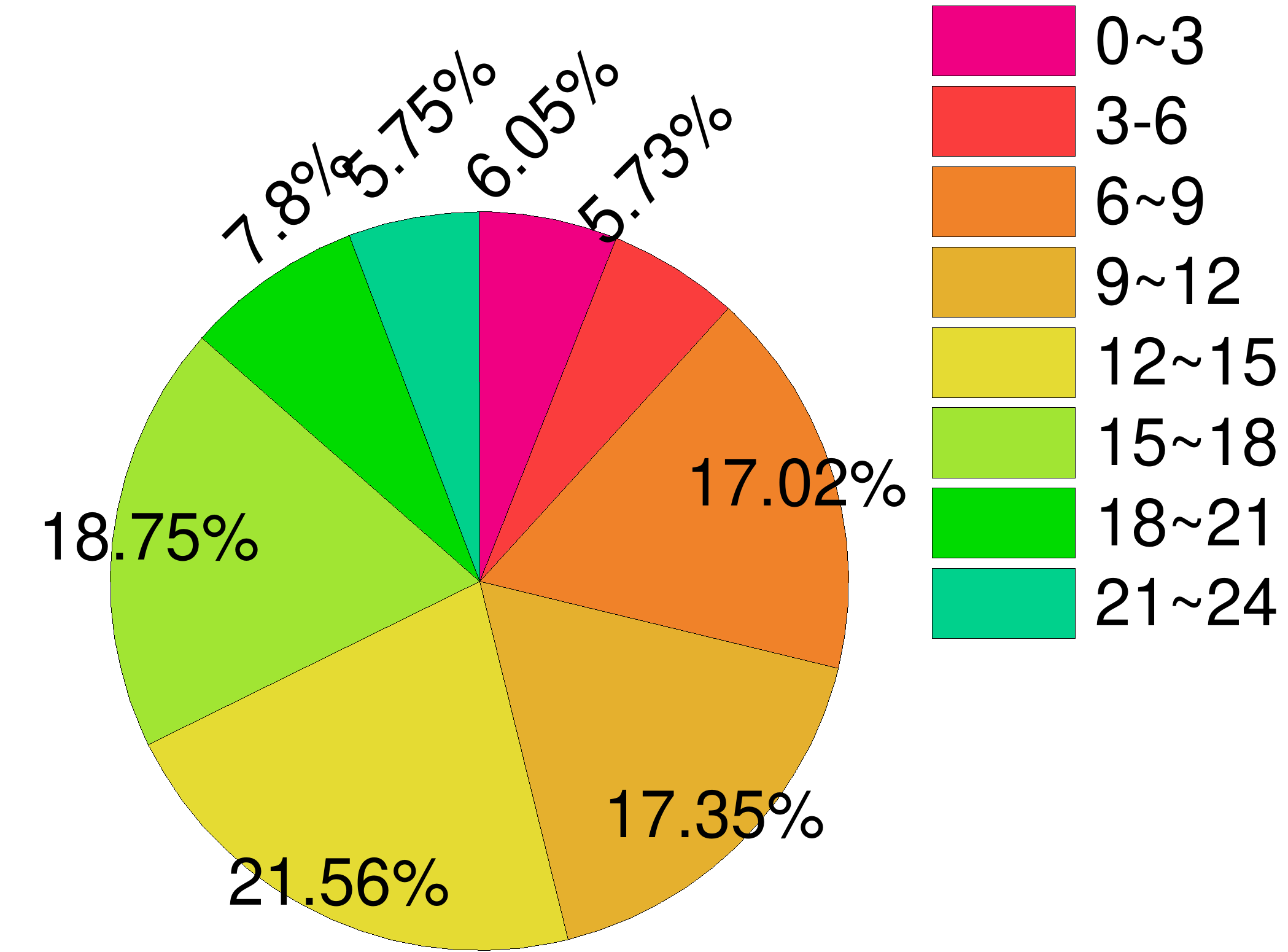}
		} 
		\subfigure[Weather condition distribution.] { \label{Fig-sector-weather} 
			\includegraphics[width=0.46\columnwidth]{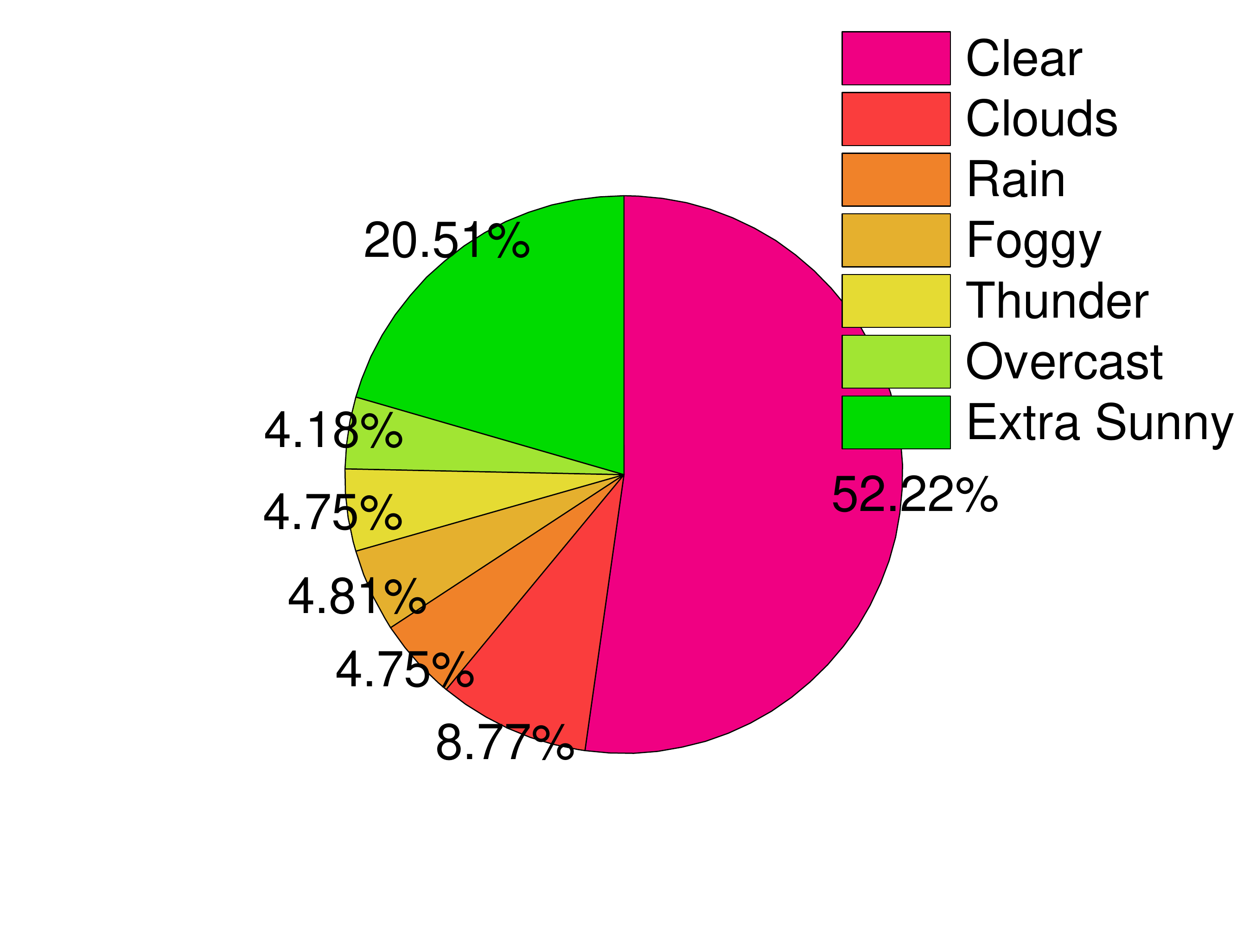}
		} 
		\caption{The pie charts of time stamp and weather condition distribution on GCC dataset. In the left pie chart, the label ``$0 \sim 3$'' denotes  the time period during $\left[ {0:00,\,\left. 3:00 \right)} \right.$ in 24 hours a day.} 
		\label{Fig-sector} 
		\vspace{-0.25cm}
	\end{figure}
	
	
	Existing datasets only focus on one of sparse or congested crowd. However, a large scene may also contain very few people in the wild. Considering that, during the generation process of an image, the number of people is set as random value in the range of its level. Therefore, GCC has more large-range than other real datasets. 
	
	\textbf{Diverse Environments.} In order to construct the data that are close to the wild, the images are captured at a random time in a day and under a random weather conditions. In GTA5, we select seven types of weathers: clear, clouds, rain, foggy, thunder, overcast and extra sunny. The last two rows of Fig. \ref{Fig-propery} illustrate the exemplars at different times and under various weathers. In the process of generation, we tend to produce more images under common conditions. The two sector charts in Fig. \ref{Fig-sector} respectively show the proportional distribution on the time stamp and weather conditions of GCC dataset.

	\section{Supervised Crowd Counting}
	FCN-based methods \cite{zhang2016single,marsden2016fully,zeng2017multi,li2018csrnet} attain good performances for crowd counting. In this section, we design an effective spatial Fully Convolutional Network (SFCN) to directly regress the density map, which is able to encode the global context information.
	
	\subsection{Network Architecture}
	
	Fully convolutional network (FCN) is proposed by Long \emph{et al.} \cite{long2015fully} in 2016, which focuses on pixel-wise task (such as semantic segmentation, saliency detection). FCN uses the convolutional layer to replace the fully connected layer in traditional CNN, which guarantees that the network can receive the image with an arbitrary size and produce the output of the corresponding size. For encoding the context information, Pan \emph{et al.} \cite{pan2017spatial} present a spatial encoder via a sequence of convolution on the four directions (down, up, left-to-right and right-to-left). 
	
	\begin{figure}
		\centering
		\includegraphics[width=0.48\textwidth]{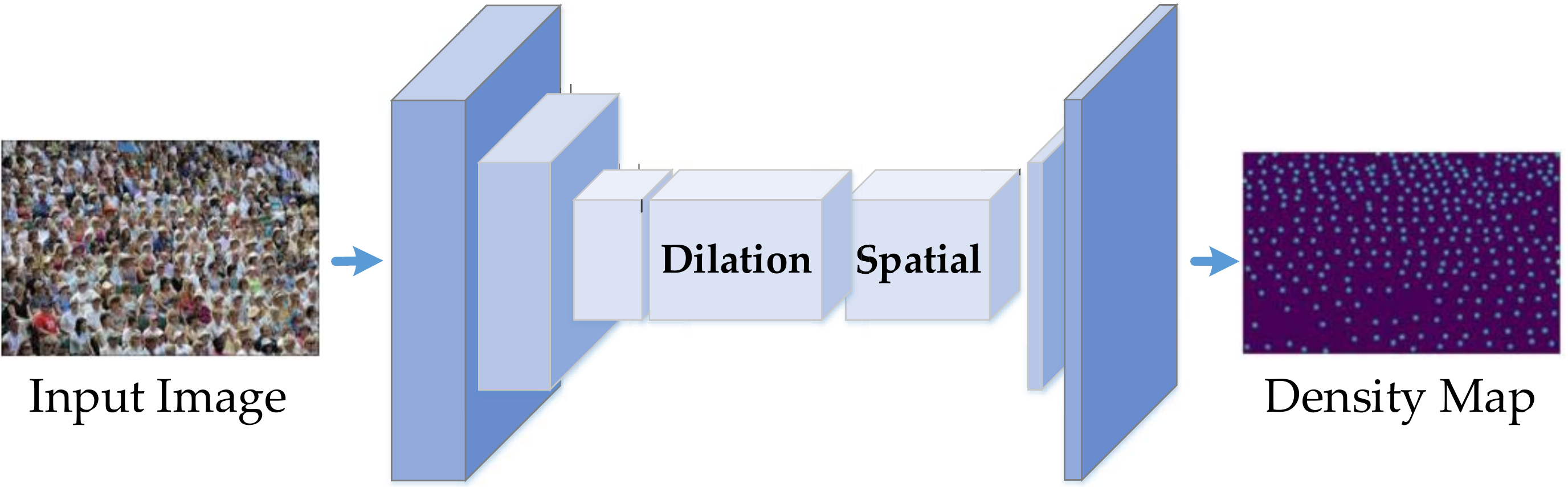}
		\caption{The architecture of spatial FCN (SFCN). }\label{Fig-sfcn}
		\vspace{-0.25cm}
	\end{figure}
	
	In this paper, we design a spatial FCN (SFCN) to produce the density map, which adopt  VGG-16 \cite{simonyan2014very} or ResnNet-101 \cite{he2016deep} as the backbone. To be specific, the spatial encoder is added to the top of the backbone. The feature map flow is illustrated as in Fig. \ref{Fig-sfcn}. After the spatial encoder, a regression layer is added, which directly outputs the density map with input's $1/8$ size. Here, we do not review the spatial encoder because of the limited space. During the training phase, the objective is minimizing standard Mean Squared Error at the pixel-wise level; the learning rate is set as ${10^{ - 5}}$; and Adam algorithm is used to optimize SFCN. 
	

	\subsection{Experiments}
	
	In this section, the two types of experiments are conducted: 1) training and testing within GCC dataset; 2) pre-training on GCC and fine-tuning on the real datasets.
	
	\subsubsection{Experiments on GCC Dataset}

	\label{trainGCC}
	
	\begin{table*}[htbp]
		\vspace{-0.05cm}
		\centering
		\caption{The results of our proposed SFCN and the three classic methods on GCC dataset.}
		\small
		\begin{tabular}{cIcc|ccIcc|ccIcc|cc}
			\whline
			\multirow{2}{*}{Method} & \multicolumn{4}{cI}{\small{Random splitting}} & \multicolumn{4}{cI}{\small{Cross-camera splitting}} & \multicolumn{4}{c}{\small{Cross-location splitting}}\\
			\cline{2-13}
			&MAE &MSE &PSNR &SSIM  &MAE &MSE &PSNR &SSIM &MAE &MSE &PSNR &SSIM \\
			\whline
			MCNN \cite{zhang2016single}   &100.9 &217.6 &24.00 &0.838  &110.0 &221.5 &23.81 &0.842  &154.8 &340.7 &24.05 &0.857\\
			\hline
			CSR \cite{li2018csrnet}  &38.2 &87.6 &29.52 &0.829 &61.1 &134.9 &29.03 &0.826  &92.2 &220.1 &28.75 &0.842\\
			\whline
			FCN &42.3 &98.7 &30.10 &0.889 &61.5 &156.6 &28.92 &0.874  &97.5 &226.8 &29.33 &0.866\\
			\hline	
			SFCN &\textbf{36.2} &\textbf{81.1} &\textbf{30.21} &\textbf{0.904} &\textbf{56.0} &\textbf{129.7} &\textbf{29.17} &\textbf{0.889} &\textbf{89.3} &\textbf{216.8} &\textbf{29.50} &\textbf{0.906}\\
			\whline
		\end{tabular}\label{Table-spvsd}
	\end{table*}
	
	We report the results of the extensive experiments within GCC dataset, which verifies SFCN from three different training strategies: random, cross-camera and cross-location splitting. To be specific, the three strategies are explained as follows. 1) \textbf{Random splitting}: the entire dataset is randomly divided into two groups as the training set (75\%) and testing set (25\%), respectively. 2) \textbf{Cross-camera splitting}: as for a specific location, one surveillance camera is randomly selected for testing and the others for training.  3) \textbf{Cross-location splitting}: we randomly choose 75/25 locations for training/testing. These scheme can effectively evaluated the algorithm on GCC. Table \ref{Table-spvsd} reports the performance of our SFCN and two popular methods (MCNN \cite{zhang2016single} and CSRNet\cite{li2018csrnet}) on the proposed GCC dataset. 
	
	
	

	\subsubsection{Experiments of Pretraining \& Finetuning}
	
	Many current  methods suffer from the over-fitting because of scarce real labeled data. Some methods (\cite{babu2018divide,shi2018crowd,idrees2018composition}) exploit the pre-trained model based on ImageNet Database \cite{deng2009imagenet}. However, the trained classification models (VGG \cite{simonyan2014very}, ResNet \cite{he2016deep} and DenseNet \cite{huang2017densely}) are not a best initialization for the regression problem: the regression layers and the specific modules are still initialized at the random or regular distributions. 
	
	In this paper, we propose a new scheme to remedy the above problems: firstly, the designed model is pretrained on the large-scale GCC Dataset; then the model pre-trained on GCC is finetuned using the real dataset. In the last step, the overall parameters are trained, which is better than traditional methods. To verify our strategy, we conduct the MCNN, CSR and SFCN on the two datasets (UCF-QNRF and SHT B). Note that SFCN adopts VGG-16 as backbone, and SFCN$\dag$ uses the ResNet101 backbone. Table \ref{Table-ft} shows the results of the comparison experiments. From it, we find that using the pretrained GCC models is better than not using or using ImageNet classification models. To be specific, for MCNN from scratch, our strategy can reduce by around 30\% estimation errors. For the SFCN using pretrained ImageNet classification model, our scheme also decrease by an average 12\% errors in four groups of experiments.
	

	\begin{table}[htbp]
		\vspace{-0.15cm}
		\centering
		\caption{The effect of pretrained GCC model on finetuning real dataset (MAE/MSE). ``*'' denotes other researchers' results.}
		\small
		\setlength{\tabcolsep}{0.7mm}{
			\begin{tabular}{c|c|c|c}
				\whline
				Method &	PreTr &UCF-QNRF &SHHT B\\
				\whline
				MCNN* &None &277/426 \cite{idrees2018composition} &26.4/41.3 \cite{zhang2016single} \\
				\hline
				MCNN  &None &281.2/445.0 &26.3/39.5 \\
				\hline
				MCNN  &GCC &\underline{199.8}/\underline{311.2}(\textcolor{red}{$\downarrow 29/30\%$}) &\underline{18.8}/\underline{28.2}(\textcolor{red}{$\downarrow 29/29\%$}) \\
				\whline
				CSR* &ImgNt &- &10.6/16.0 \cite{li2018csrnet} \\
				\hline
				CSR  &ImgNt &120.3/208.5 &10.6/16.6 \\
				\hline
				CSR  &GCC &\underline{112.4}/\underline{185.6}(\textcolor{red}{$\downarrow 7/11\%$}) &\underline{10.1}/\underline{15.7}(\textcolor{red}{$\downarrow 5/5\%$}) \\
				\whline
				SFCN  &ImgNt &134.3/240.3 &11.0/17.1  \\
				\hline
				SFCN  &GCC &\underline{124.7}/\underline{203.5}(\textcolor{red}{$\downarrow 7/15\%$}) &\underline{9.4}/\underline{14.4}(\textcolor{red}{$\downarrow 15/16\%$})  \\
				\whline
				SFCN\dag &ImgNt &114.8/192.0 & 8.9/14.3 \\
				\hline
				SFCN\dag  &GCC &\underline{\textbf{102.0}}/\underline{\textbf{171.4}}(\textcolor{red}{$\downarrow 11/11\%$}) &\underline{\textbf{7.6}}/\underline{\textbf{13.0}}(\textcolor{red}{$\downarrow 15/9\%$})  \\
				\whline
			\end{tabular}
		}
		\vspace{-0.25cm}
		\label{Table-ft}
	\end{table}
	
	We also present the final results of our SFCN$\dag$ on five real datasets, which is fintuned on the pretrained SFCN$\dag$ using GCC. Compared with the state-of-the-art performance, SFCN$\dag$ refreshes the records on the four datasets. The detailed results comparison is listed in the Table \ref{Table-soa}.
	\begin{table}[htbp]
		\vspace{-0.15cm}
		\centering
		\caption{The comparison with the state-of-the-art performance on real datasets.}
		\small
		\begin{tabular}{c|c|c}
			\whline
			\multirow{2}{*}{Dataset} &\multicolumn{2}{c}{Results (MAE/MSE)}\\
			\cline{2-3} &SOTA &SFCN\dag \\
			\whline
			UCF-QNRF \cite{idrees2018composition} & CL\cite{idrees2018composition}: 132/191 &\textbf{102.0}/\textbf{171.4}  \\
			\hline
			SHT A \cite{zhang2016single}  &SA\cite{cao2018scale}: 67.0/\textbf{104.5} &\textbf{64.8}/107.5\\
			\hline
			SHT B \cite{zhang2016single} &SA\cite{cao2018scale}: 8.4/13.6 & \textbf{7.6}/\textbf{13.0}\\
			\hline
			UCF\_CC\_50 \cite{idrees2013multi}  &SAN\cite{liu2018crowd}:219.2/\textbf{250.2} &\textbf{214.2}/318.2  \\
			\hline
			WorldExpo'10\cite{zhang2016data}   &ACSCP\cite{shen2018crowd}:\textbf{7.5}(MAE) & 9.4(MAE) \\
			\whline
		\end{tabular}
		\label{Table-soa}
		\vspace{-0.5cm}
	\end{table}

	\section{Crowd Counting via Domain Adaptation}
	
	\label{DA}
	
	\begin{figure*}
		\centering
		\includegraphics[width=0.98\textwidth]{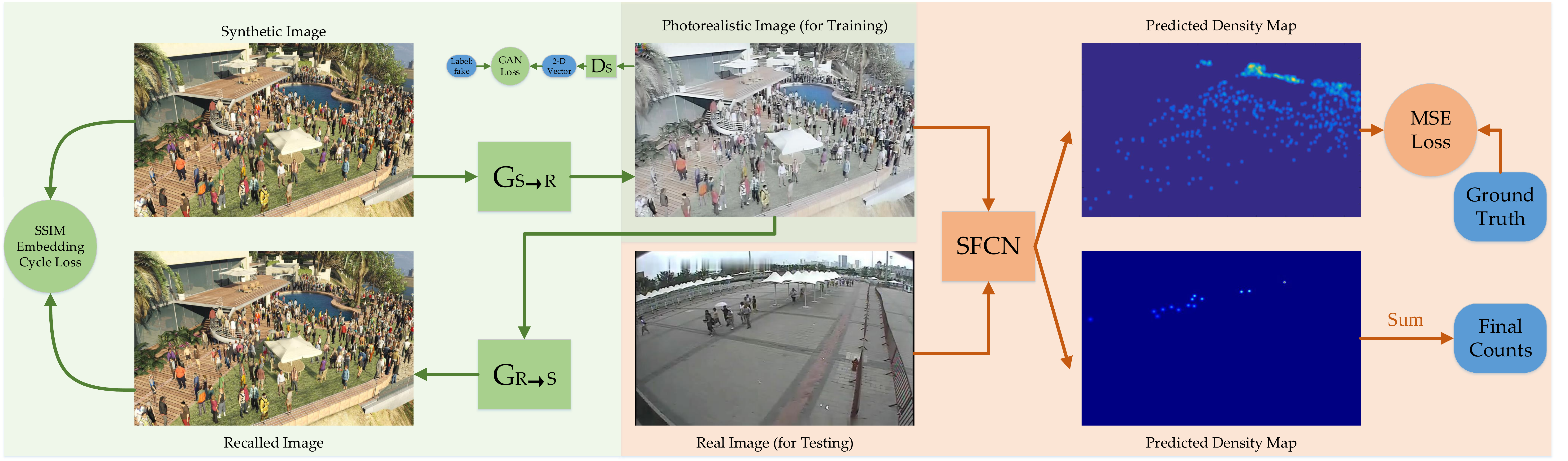}
		\caption{The flowchart of the proposed crowd counting via domain adaptation. The light green region is SSIM Embedding (SE) Cycle GAN, and light orange region represents Spatial FCN (FCN). Limited by paper length, we do not show the adaptation flowchart of real images to synthetic images (R$\rightarrow$S), which is similar to S$\rightarrow$R.}\label{Fig-cyclegan}
		\vspace{-0.25cm}
	\end{figure*}
	
	The last section proposes the supervised learning on synthetic or real datasets, which adopts the labels of real data. For extremely congested scenes, manually annotating them is a tedious work. Not only that, there are label errors in man-made annotations. Therefore, we attempt to propose a crowd counting method via domain adaptation to save manpower, which learns specific patterns or features from the synthetic data and transfers them to the real world. Through this thought, we do not need any manual labels of real data. Unfortunately, the generated synthetic data are very different from real-world data (such as in color style, texture and so on), which is treated as ``domain gap''. Even in real life, the domain gap is also very common. For example, Shanghai Tech Part B and WorldExpo'10 are captured in different locations from different cameras, which causes that the data of them are quite different. Thus, it is an important task that how to transfer effective features between different domains, which is named as a ``Domain Adaptation'' (DA) problem.
	
	In this work, we propose a crowd counting method via domain adaptation, which can effectively learn domain-invariant feature between synthetic and real data. To be specific, we present a SSIM Embedding (SE) Cycle GAN to transform the synthetic image to the photo-realistic image. Then we will train a SFCN on the translated data. Finally, we directly test the model on the real data. The entire process does not need any manually labeled data.  Fig. \ref{Fig-cyclegan} demonstrates the flowchart of the proposed method.

	\subsection{SSIM Embedding Cycle GAN}
	
	Here, we recall the crowd counting via domain adaptation by mathematical notations. The purpose of DA is to learn translation mapping between the synthetic domain ${\mathcal{S}}$ and the real-world domain ${\mathcal{R}}$. The synthetic domain ${\mathcal{S}}$ provides images ${I_\mathcal{S}}$ and count labels $L_\mathcal{S}$. And the real-world domain ${\mathcal{R}}$ only provides images ${I_\mathcal{R}}$. In a word, given ${{\rm{i}}_\mathcal{S} \in I_\mathcal{S}}$,  ${\rm{l}}_\mathcal{S} \in L_\mathcal{S}$ and ${{\rm{i}}_\mathcal{R} \in I_\mathcal{R}}$ (the lowercase letters represent the samples in the corresponding sets), we want to train a crowd counter to predict density maps of ${\mathcal{R}}$.

	\textbf{Cycle GAN.} The original Cycle GAN \cite{zhu2017unpaired} is proposed by Zhu \emph{et al.}, which focuses on unpaired image-to-image translation. For different two domains, we can exploit Cycle GAN to handle DA problem, which can translate the synthetic images to photo-realistic images. 
	As for the domain ${\mathcal{S}}$ and ${\mathcal{R}}$, we define two generator ${G_{\mathcal{S}\rightarrow\mathcal{R}}}$ and ${G_{\mathcal{R}\rightarrow\mathcal{S}}}$. The former one attempts to learn a mapping function from domain ${\mathcal{S}}$ to ${\mathcal{R}}$, and vice versa, the latter one's goal is to learn the mapping from domain ${\mathcal{R}}$ to ${\mathcal{S}}$. Following \cite{zhu2017unpaired}, we introduce the cycle-consistent loss to regularize the training process. To be specific, for the sample ${\rm{i}}_\mathcal{S}$ and ${\rm{i}}_\mathcal{R}$, one of our objective is ${\rm{i}}_\mathcal{S} \rightarrow {G_{\mathcal{S}\rightarrow\mathcal{R}}}({\rm{i}}_\mathcal{S})\rightarrow {G_{\mathcal{R}\rightarrow\mathcal{S}}}({G_{\mathcal{S}\rightarrow\mathcal{R}}}({\rm{i}}_\mathcal{S}))\approx {\rm{i}}_\mathcal{S}$. Another objective is inverse process for ${\rm{i}}_\mathcal{R}$. The cycle-consistent loss is an L1 penalty in the cycle architecture, which is defined as follows:
	\begin{equation}
	\begin{array}{l}
	\begin{aligned}
	{\mathcal{L}_{cycle}}( & {G_{\mathcal{S}\rightarrow \mathcal{R}}},{G_{\mathcal{R}\rightarrow\mathcal{S}}},\mathcal{S},\mathcal{R})\\
	& = {\mathbb{E}_{{i_\mathcal{S}} \sim {I_\mathcal{S}}}}[{\left\| {{G_{\mathcal{R}\rightarrow\mathcal{S}}}({G_{\mathcal{S}\rightarrow{R}}}({i_\mathcal{S}})) - {i_\mathcal{S}}} \right\|_1}]\\
	& + {\mathbb{E}_{{i_\mathcal{R}} \sim {I_\mathcal{R}}}}[{\left\| {{G_{\mathcal{S}\rightarrow\mathcal{R}}}({G_{\mathcal{R}\rightarrow{S}}}({i_\mathcal{R}})) - {i_\mathcal{R}}} \right\|_1}].
	\end{aligned}
	\end{array}
	\end{equation}
	
	Additionally, two discriminators ${D_{\mathcal{R}}}$ and ${D_{\mathcal{S}}}$ are modeled corresponding to the ${G_{\mathcal{S}\rightarrow\mathcal{R}}}$ and ${G_{\mathcal{R}\rightarrow\mathcal{S}}}$. Specifically, ${D_{\mathcal{R}}}$ attempts to discriminate that where the images are from ( $I_\mathcal{R}$ or ${G_{\mathcal{S}\rightarrow\mathcal{R}}}(I_\mathcal{S})$), and ${D_{\mathcal{S}}}$ tries to discriminate the images from $I_\mathcal{S}$ or ${G_{\mathcal{R}\rightarrow\mathcal{S}}}(I_\mathcal{R})$. Take ${D_{\mathcal{R}}}$ for example, and the training objective is adversarial loss \cite{goodfellow2014generative}, which is formulated as:
	\begin{equation}
	\begin{array}{l}
	\begin{aligned}
	{\mathcal{L}_{GAN}}(& {G_{\mathcal{S}\rightarrow \mathcal{R}}},{D_\mathcal{R}},\mathcal{S},\mathcal{R}) \\
	& = {\mathbb{E}_{{i_\mathcal{R}} \sim {I_\mathcal{R}}}}[log({D_\mathcal{R}}({i_\mathcal{R}})] \\
	& + {\mathbb{E}_{{i_\mathcal{S}} \sim {I_\mathcal{S}}}}[log(1 - {D_\mathcal{R}}({G_{\mathcal{S}\rightarrow \mathcal{R}}}({i_\mathcal{S}}))].
	\end{aligned}
	\end{array}
	\end{equation}
	
	The final loss function is defined as:
	\begin{equation}
	\begin{array}{l}
	\begin{aligned}
	{\mathcal{L}_{CycleGAN}}(&{G_{\mathcal{S}\rightarrow \mathcal{R}}},{G_{\mathcal{R}\rightarrow \mathcal{S}}}, {D_\mathcal{R}},{D_\mathcal{S}},\mathcal{S},\mathcal{R}) \\
	& = {\mathcal{L}_{GAN}}({G_{\mathcal{S}\rightarrow \mathcal{R}}},{D_\mathcal{R}},\mathcal{S},\mathcal{R})\\
	& + {\mathcal{L}_{GAN}}({G_{\mathcal{R}\rightarrow \mathcal{S}}},{D_\mathcal{S}},\mathcal{S},\mathcal{R}) \\
	& + \lambda {\mathcal{L}_{cycle}}({G_{\mathcal{S}\rightarrow \mathcal{R}}},{G_{\mathcal{R}\rightarrow\mathcal{S}}},\mathcal{S},\mathcal{R}),
	\end{aligned}
	\end{array}
	\end{equation}
	where $\lambda$ is the weight of cycle-consistent loss. 
	
	\begin{table*}[htbp]
		\centering
		\small
		\caption{The performance of no adaptation (No Adpt), Cycle GAN and SE Cycle GAN (ours) on the five real-world datasets.}
		\setlength{\tabcolsep}{1.60mm}{
			\begin{tabular}{c|cIc|c|c|cIc|c|c|cIc|c|c|c}
				\hline
				\multirow{2}{*}{Method}	&\multirow{2}{*}{DA} &\multicolumn{4}{cI}{SHT A} &\multicolumn{4}{cI}{SHT B} &\multicolumn{4}{c}{UCF\_CC\_50}\\
				\cline{3-14} 
				& & MAE &MSE &PSNR &SSIM &MAE & MSE &PSNR &SSIM  &MAE & MSE &PSNR &SSIM\\
				\hline
				NoAdpt  &\xmark &160.0 &216.5 &19.01 &0.359 &22.8 &30.6 &24.66 &0.715 &487.2 &689.0 &17.27 &0.386 \\
				\hline
				Cycle GAN\cite{zhu2017unpaired} &\rmark &143.3 &204.3 &\textbf{19.27} &0.379  &25.4 &39.7 &24.60 &0.763  &404.6 &548.2 &\textbf{17.34} &0.468 \\
				\hline	
				SE Cycle GAN (ours)  &\rmark&\textbf{123.4} &\textbf{193.4} &18.61 &\textbf{0.407} &\textbf{19.9} &\textbf{28.3} &\textbf{24.78} &\textbf{0.765} &\textbf{373.4} &\textbf{528.8} &17.01 &\textbf{0.743} \\
				\hline
			\end{tabular}
		}
		\setlength{\tabcolsep}{2.0mm}{
			\begin{tabular}{c|cIc|c|c|cIc|c|c|c|c|c}
				\hline
				\multirow{2}{*}{Method}	&\multirow{2}{*}{DA} &\multicolumn{4}{cI}{UCF-QNRF} &\multicolumn{6}{c}{WorldExpo'10 (MAE)}\\
				\cline{3-12} 
				& & MAE &MSE &PSNR &SSIM &S1 &S2 &S3 &S4 &S5 &Avg. \\
				\hline
				
				NoAdpt  &\xmark &275.5 &458.5 &20.12 &0.554 &4.4 &87.2 &59.1  &51.8 &11.7 &42.8 \\
				\hline
				Cycle GAN\cite{zhu2017unpaired} &\rmark &257.3 &400.6 &20.80 &0.480  &4.4 &69.6 &49.9 &29.2  &9.0 &32.4   \\
				\hline	
				SE Cycle GAN (ours)  &\rmark&\textbf{230.4} &\textbf{384.5} &\textbf{21.03} &\textbf{0.660} &\textbf{4.3} &\textbf{59.1} &\textbf{43.7}  &\textbf{17.0} &\textbf{7.6} &\textbf{26.3}  \\
				\hline
			\end{tabular}
		}
		\label{Table-DA}
		\vspace{-0.5cm}
	\end{table*}
	
	\textbf{SSIM Embedding Cycle-consistent loss.} In the crowd scenes, the biggest differences between high-density regions and other regions (low-density regions or background) is the local patterns and texture features. 
	Unfortunately, in the translation from synthetic to real images, the original cycle consistency is prone to losing them, which causes that the translated images lose the detailed information and are easily distorted. 
	
	To remedy the aforementioned problem, we introduce Structural Similarity Index (SSIM) \cite{wang2004image} into the traditional CycleGAN, which is named as ``SE Cycle GAN''. SSIM is an indicator widely used in the field of image quality assessment, which computes the similarity between two images in terms of local patterns (mean, variance and covariance). About the SSIM in crowd counting, CP-CNN \cite{sindagi2017generating} is the first to evaluate the density map using SSIM, and SANet \cite{cao2018scale} adopt SSIM loss to generate high-quality density maps. 
	
	Similar to the traditional cycle consistency, our goal is:  ${G_{\mathcal{R}\rightarrow\mathcal{S}}}({G_{\mathcal{S}\rightarrow\mathcal{R}}}({\rm{i}}_\mathcal{S}))\approx {\rm{i}}_\mathcal{S}$. To be specific, in addition to L1 penalty, the SSIM penalty is added to the training process. The range of SSIM value is in $\left[ { - 1,1} \right]$, and larger SSIM means that the image has more higher quality. In particular, when the two images are identical, the SSIM value is equal to $1$. In the practice, we convert the SSIM value into the trainable form, which is defined as:

	\begin{equation}
	\label{eq11}
	\begin{split}
	& {\mathcal{L}_{SEcycle}}({G_{\mathcal{S}\rightarrow \mathcal{R}}},{G_{\mathcal{R}\rightarrow\mathcal{S}}},\mathcal{S},\mathcal{R})\\
	& = {\mathbb{E}_{{i_\mathcal{S}} \sim {I_\mathcal{S}}}}[ {1-SSIM({i_\mathcal{S}},{G_{\mathcal{R}\rightarrow\mathcal{S}}}({G_{\mathcal{S}\rightarrow{R}}}({i_\mathcal{S}}))})]\\
	& + {\mathbb{E}_{{i_\mathcal{R}} \sim {I_\mathcal{R}}}}[ {1-SSIM({i_\mathcal{R}},{G_{\mathcal{S}\rightarrow\mathcal{R}}}({G_{\mathcal{R}\rightarrow{S}}}({i_\mathcal{R}}))})],
	\end{split}
	\end{equation}
	where $SSIM( \cdot , \cdot )$ is standard computation: the parameter settings are directly followed by \cite{wang2004image}. The first input is the original image from domain $\mathcal{S}$ or $\mathcal{R}$, and the second input is the reconstructed image produced by the two generators in turns. Finally, the final objective of SE Cycle GAN is defined as:
	\begin{equation}
	\begin{array}{l}
	\begin{aligned}
	{\mathcal{L}_{ours}}(&{G_{\mathcal{S}\rightarrow \mathcal{R}}},{G_{\mathcal{R}\rightarrow \mathcal{S}}}, {D_\mathcal{R}},{D_\mathcal{S}},\mathcal{S},\mathcal{R}) \\
	& = {\mathcal{L}_{GAN}}({G_{\mathcal{S}\rightarrow \mathcal{R}}},{D_\mathcal{R}},\mathcal{S},\mathcal{R})\\
	& + {\mathcal{L}_{GAN}}({G_{\mathcal{R}\rightarrow \mathcal{S}}},{D_\mathcal{S}},\mathcal{S},\mathcal{R}) \\
	& + \lambda {\mathcal{L}_{cycle}}({G_{\mathcal{S}\rightarrow \mathcal{R}}},{G_{\mathcal{R}\rightarrow\mathcal{S}}},\mathcal{S},\mathcal{R}) \\
	& + \mu {\mathcal{L}_{SEcycle}}({G_{\mathcal{S}\rightarrow \mathcal{R}}},{G_{\mathcal{R}\rightarrow\mathcal{S}}},\mathcal{S},\mathcal{R}),
	\end{aligned}
	\end{array}
	\end{equation}
	where $\lambda$ and $\mu$ are the weights of cycle-consistent and SSIM Embedding cycle-consistent loss, respectively. During the training phase, the $\mu$ is set as $1$, other parameters and settings are the same as Cycle GAN \cite{zhu2017unpaired}. 
	
	\textbf{Density/Scene Regularization.} For a better domain adaptation from synthetic to real world, we design two strategies to facilitate the DA model to learn domain-invariant feature and produce the valid density map.\label{SDR}
	
	Although we translate synthetic images to photo-realistic images, some objects and data distributions in the real world are unseen during training the translated images. As a pixel-wise regression problem, the density may be an arbitrary value in theory. In fact, in some preliminary experiments, we find some backgrounds in real data are estimated as some exceptionally large values. To handle this problem, we set a upper bound $MAX_{\mathcal{S}}$, which is defined as the max density in the synthetic data. If the output value of a pixel is more than $MAX_{\mathcal{S}}$, the output will be set as $0$. Note that the network's last layer is ReLU, so the output of each pixel must be greater than or equal to $0$. 
	
	Since GCC is large-counter-range and diverse dataset, using all images may cause the side effect in domain adaptation. For example, ShanghaiTech does not contain the thunder/rain scenes, and WorldExpo'10 does not have the scene that can accommodate more than 500 people. Training all translated synthetic images can decrease the adaptation performance on the specific dataset. Thus, we manually select some specific scenes for different datasets. The concrete strategies are described in the supplementary. In general, it is a coarse data filter not an elaborate selection.


	\subsection{Experiments}
	

	\begin{figure*}
		\centering
		\includegraphics[width=0.98\textwidth]{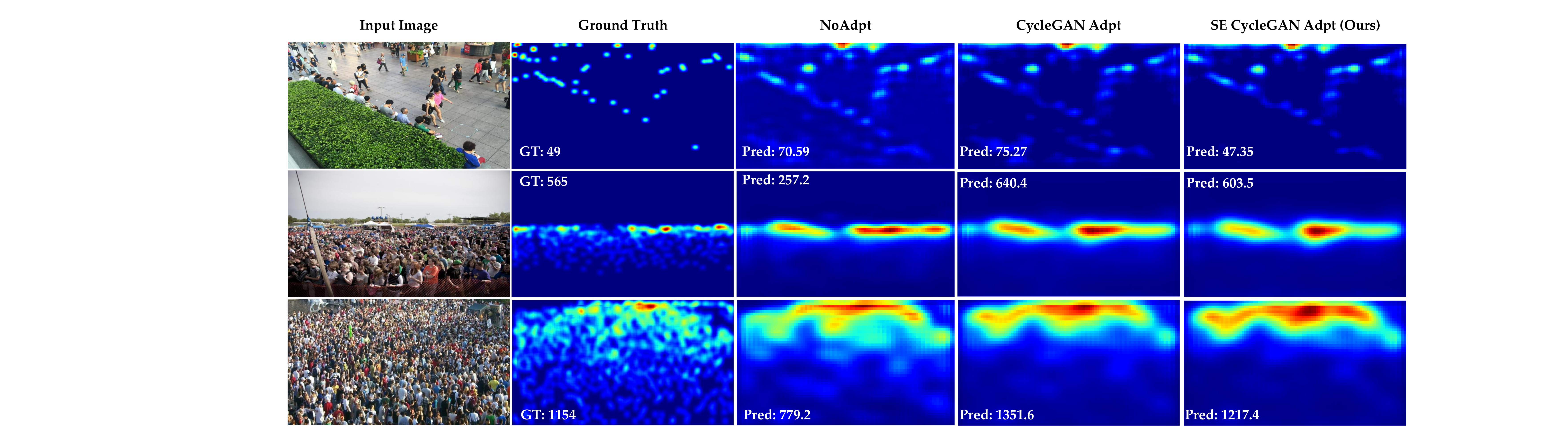}
		\caption{The demonstration of different methods on SHT dataset. ``GT'' and ``Pred'' represent the labeled and predicted count, respectively. }\label{Fig-dare}
		\vspace{-0.25cm}
	\end{figure*}
	
	\subsubsection{Performance on Real-world Datasets}
	
	In this section, we conduct the adaptation experiments from GCC dataset to five mainstream real-world datasets: ShanghaiTech A/B \cite{zhang2016single}, UCF\_CC\_50 \cite{idrees2013multi}, UCF-QNRF \cite{idrees2018composition} and WorldExpo'10\cite{zhang2016data}. For the best performance, all models adopt the Scene/Density Regularization mentioned in Section \ref{SDR}. 

	Table \ref{Table-DA} shows the results of the No Adaptation (No Adpt), Cycle GAN and the proposed SSIM Embedding (SE) Cycle GAN. From it, we find the results after adaptation are far better than that of no adaptation, which indicates the adaptation can effectively reduce the domain gaps between synthetic and real-world data. After embedding SSIM loss in cycle GAN, almost all performances are improved on five datasets. There are only two reductions of PSNR on Shanghai Tech A and UCF\_CC\_50. In general, the proposed SE Cycle GAN outperforms the original Cycle GAN. In addition, we find the results on Shanghai Tech B achieve a good level, even outperforms some early supervised methods \cite{zhang2016single, sindagi2017cnn, sam2017switching, sindagi2017generating,liu2018decidenet}. The main reasons are: 1) the real data is strongly consistent, which is captured by the same sensors; 2) the data has high image clarity. The two characteristics guarantee that the SE CycleGAN's adaptation on Shanghai Tech B is more effective than others.
	
	Fig. \ref{Fig-dare} demonstrates three groups of visualized results on Shanghai Tech dataset. Compared with no adaptation, the map quality via Cycle GAN has a significant improvement. From Row 1, we find the predicted maps are very close to the groundtruth.  However, for the extremely congested scenes (in Row 2 and 3), the results are far from the ground truth. We think the main reason is that the translated images lose the details (such as texture, sharpness and edge) in high-density regions.

	\subsubsection{Analysis of SE \& DSR}
	
	\begin{figure}
		\centering
		\includegraphics[width=0.45\textwidth]{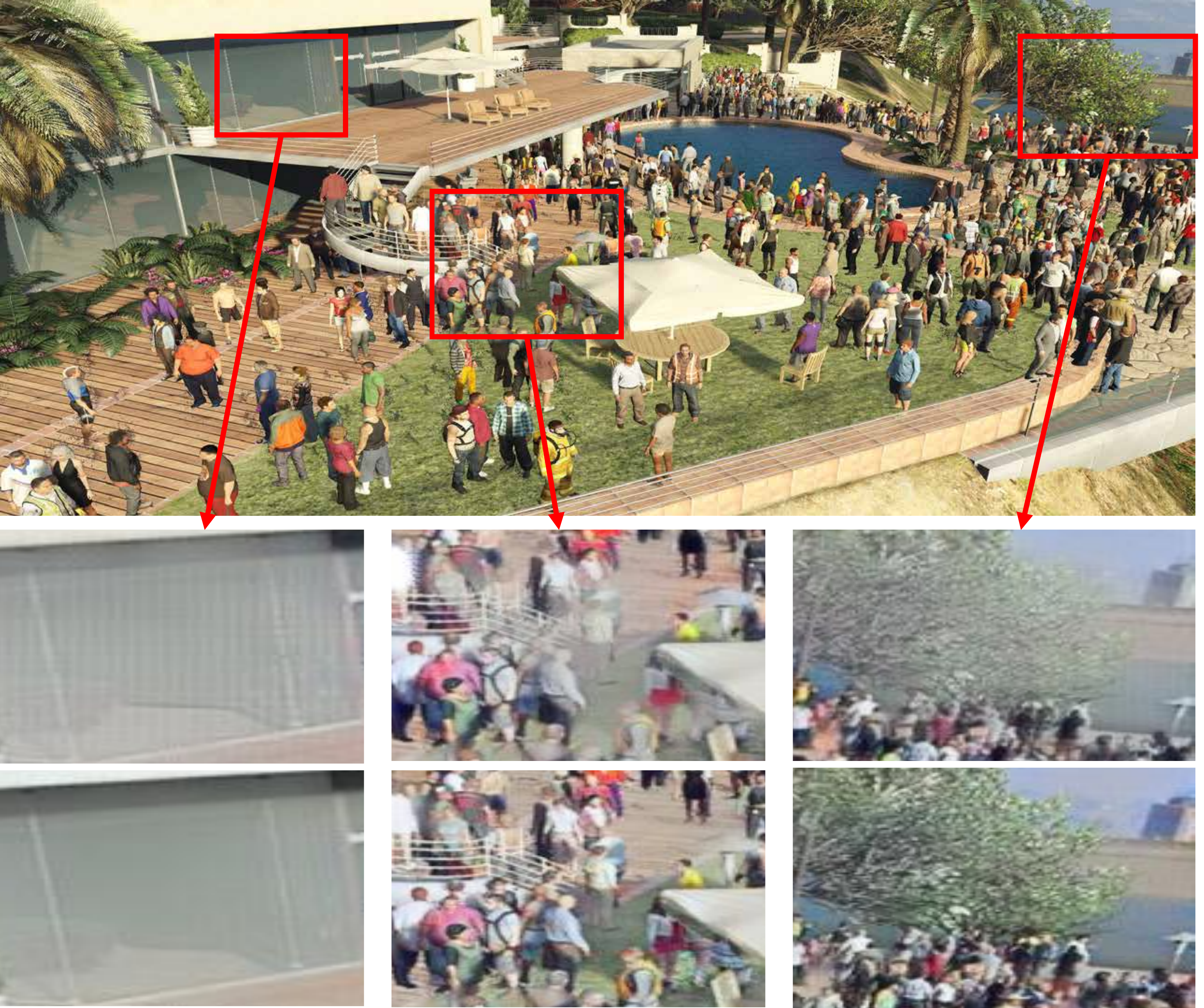}
		\caption{The comparison of Cycle GAN and SE Cycle GAN. }\label{Fig-ssim}
		\vspace{-0.25cm}
	\end{figure}
	
	\textbf{SSIM Embedding. } SSIM Embedding can guarantee the original synthetic and reconstructed images have high structural similarity(SS), which prompts two generators' translation for images maintain a certain degree of SS during the training process. Fig. \ref{Fig-ssim} illustrates the visualizations of two adaptations, where the first row is original images, the second and third row are translated images of Cycle GAN and SE Cycle GAN. Through comparison, the latter is able to retain local texture and structural similarity.

	\textbf{Density/Scene Regularization.} Here, we compare the performance of three model (No Adpt, Cycle GAN and SE Cycle GAN) without Density/Scene Regularization (DSR) and with DSR. Table \ref{Table-dsr} reports the performance of with or without DSR on SHT A dataset. From the results in first column, we find these two adaptation methods cause some side effects. In fact, they do not produce the ideal translated images. When introducing DSR, the nonexistent synthetic scenes in the real datasets are filtered out, which improves the domain adaptation performance.
	
	\begin{table}[htbp]
		
		\centering
		\caption{The results under different configurations on SHT A.}
		\small
		\begin{tabular}{c|c|c}
			\hline
			Method	&w/o DSR &with DSR\\
			\hline
			NoAdpt &\textbf{163.6}/244.5 &\underline{160.0}/\underline{216.5}   \\
			\hline
			Cycle GAN \cite{zhu2017unpaired} &180.1/290.3 &\underline{143.3}/\underline{204.3}  \\
			\hline
			SE Cycle GAN  &169.8/\textbf{230.2} &\underline{\textbf{123.4}}/\underline{\textbf{193.4}}  \\
			\hline
		\end{tabular}
		\label{Table-dsr}
		\vspace{-0.5cm}
	\end{table}


	\section{Conclusion}
	
	In this paper, we are committed to improving the performance of crowd counting in the wild. To this end, we firstly develop an automatic data collector/labeler and construct a large-scale synthetic crowd counting dataset. Exploiting the generated data, we then propose two effective ways (supervised learning and domain adaptation) to significantly improve the counting performance in the wild. The experiments demonstrate that the supervised method achieves the state-of-the-art performance and the domain adaptation method obtains acceptable results. In the future work, we will focus on the crowd counting via domain adaptation, and further explore that how to extract more effective domain-invariant features between synthetic and real-world data. 
	
	\textbf{Acknowledgments.} This work was supported by the National Natural Science Foundation of China under Grant U1864204 and 61773316, State Key Program of National Natural Science Foundation of China under Grant 61632018, Natural Science Foundation of Shaanxi Province under Grant 2018KJXX-024, and Project of Special Zone for National Defense Science and Technology Innovation.

	{\small
		\bibliographystyle{ieee}
		\bibliography{egbib}

\begin{thebibliography}{10}\itemsep=-1pt

\bibitem{rstar}
Rockstar games.
\newblock \url{https://www.rockstargames.com/}.

\bibitem{hookv}
Script hook v.
\newblock \url{http://www.dev-c.com/gtav/scripthookv/}.

\bibitem{unity}
Unity engine.
\newblock \url{https://unity3d.com/}.

\bibitem{unreal}
Unreal engine.
\newblock \url{https://www.unrealengine.com/}.

\bibitem{babu2018divide}
D.~Babu~Sam, N.~N. Sajjan, R.~Venkatesh~Babu, and M.~Srinivasan.
\newblock Divide and grow: Capturing huge diversity in crowd images with
  incrementally growing cnn.
\newblock In {\em Proceedings of the IEEE Conference on Computer Vision and
  Pattern Recognition}, pages 3618--3626, 2018.

\bibitem{bak2018domain}
S.~Bak, P.~Carr, and J.-F. Lalonde.
\newblock Domain adaptation through synthesis for unsupervised person
  re-identification.
\newblock {\em arXiv preprint arXiv:1804.10094}, 2018.

\bibitem{cao2018scale}
X.~Cao, Z.~Wang, Y.~Zhao, and F.~Su.
\newblock Scale aggregation network for accurate and efficient crowd counting.
\newblock In {\em Proceedings of the European Conference on Computer Vision},
  pages 734--750, 2018.

\bibitem{chan2008privacy}
A.~B. Chan, Z.-S.~J. Liang, and N.~Vasconcelos.
\newblock Privacy preserving crowd monitoring: Counting people without people
  models or tracking.
\newblock In {\em Proceedings of the IEEE conference on Computer Vision and
  Pattern Recognition}, pages 1--7, 2008.

\bibitem{chen2012feature}
K.~Chen, C.~C. Loy, S.~Gong, and T.~Xiang.
\newblock Feature mining for localised crowd counting.
\newblock In {\em Proceedings of the British Machine Vision Conference},
  volume~1, page~3, 2012.

\bibitem{deng2009imagenet}
J.~Deng, W.~Dong, R.~Socher, L.-J. Li, K.~Li, and L.~Fei-Fei.
\newblock Imagenet: A large-scale hierarchical image database.
\newblock In {\em Proceedings of the IEEE conference on Computer Vision and
  Pattern Recognition}, pages 248--255, 2009.

\bibitem{goodfellow2014generative}
I.~Goodfellow, J.~Pouget-Abadie, M.~Mirza, B.~Xu, D.~Warde-Farley, S.~Ozair,
  A.~Courville, and Y.~Bengio.
\newblock Generative adversarial nets.
\newblock In {\em Proceedings of the Advances in Neural Information Processing
  Systems}, pages 2672--2680, 2014.

\bibitem{he2016deep}
K.~He, X.~Zhang, S.~Ren, and J.~Sun.
\newblock Deep residual learning for image recognition.
\newblock In {\em Proceedings of the IEEE conference on Computer Vision and
  Pattern Recognition}, pages 770--778, 2016.

\bibitem{huang2017densely}
G.~Huang, Z.~Liu, L.~Van Der~Maaten, and K.~Q. Weinberger.
\newblock Densely connected convolutional networks.
\newblock In {\em Proceedings of the IEEE conference on Computer Vision and
  Pattern Recognition}, pages 4700--4708, 2017.

\bibitem{idrees2013multi}
H.~Idrees, I.~Saleemi, C.~Seibert, and M.~Shah.
\newblock Multi-source multi-scale counting in extremely dense crowd images.
\newblock In {\em Proceedings of the IEEE conference on Computer Vision and
  Pattern Recognition}, pages 2547--2554, 2013.

\bibitem{idrees2018composition}
H.~Idrees, M.~Tayyab, K.~Athrey, D.~Zhang, S.~Al-Maadeed, N.~Rajpoot, and
  M.~Shah.
\newblock Composition loss for counting, density map estimation and
  localization in dense crowds.
\newblock {\em arXiv preprint arXiv:1808.01050}, 2018.

\bibitem{Johnson-Roberson:2017aa}
M.~Johnson-Roberson, C.~Barto, R.~Mehta, S.~N. Sridhar, K.~Rosaen, and
  R.~Vasudevan.
\newblock Driving in the matrix: Can virtual worlds replace human-generated
  annotations for real world tasks?
\newblock In {\em Proceedings of the IEEE International Conference on Robotics
  and Automation}, pages 1--8, 2017.

\bibitem{DBLP:journals/spm/JuniorMJ10}
J.~C. S.~J. Junior, S.~R. Musse, and C.~R. Jung.
\newblock Crowd analysis using computer vision techniques.
\newblock {\em Signal Processing Magazine IEEE}, 27(5):66--77, 2010.

\bibitem{DBLP:conf/aaai/LiCNW17}
X.~Li, M.~Chen, F.~Nie, and Q.~Wang.
\newblock A multiview-based parameter free framework for group detection.
\newblock In {\em Proceedings of the Thirty-First AAAI Conference on Artificial
  Intelligence}, pages 4147--4153, 2017.

\bibitem{li2018csrnet}
Y.~Li, X.~Zhang, and D.~Chen.
\newblock Csrnet: Dilated convolutional neural networks for understanding the
  highly congested scenes.
\newblock In {\em Proceedings of the IEEE Conference on Computer Vision and
  Pattern Recognition}, pages 1091--1100, 2018.

\bibitem{liu2018decidenet}
J.~Liu, C.~Gao, D.~Meng, and A.~G. Hauptmann.
\newblock Decidenet: Counting varying density crowds through attention guided
  detection and density estimation.
\newblock In {\em Proceedings of the IEEE Conference on Computer Vision and
  Pattern Recognition}, pages 5197--5206, 2018.

\bibitem{liu2018crowd}
L.~Liu, H.~Wang, G.~Li, W.~Ouyang, and L.~Lin.
\newblock Crowd counting using deep recurrent spatial-aware network.
\newblock {\em arXiv preprint arXiv:1807.00601}, 2018.

\bibitem{liu2018leveraging}
X.~Liu, J.~van~de Weijer, and A.~D. Bagdanov.
\newblock Leveraging unlabeled data for crowd counting by learning to rank.
\newblock {\em arXiv preprint arXiv:1803.03095}, 2018.

\bibitem{long2015fully}
J.~Long, E.~Shelhamer, and T.~Darrell.
\newblock Fully convolutional networks for semantic segmentation.
\newblock In {\em Proceedings of the IEEE Conference on Computer Vision and
  Pattern Recognition}, pages 3431--3440, 2015.

\bibitem{marsden2016fully}
M.~Marsden, K.~McGuinness, S.~Little, and N.~E. O'Connor.
\newblock Fully convolutional crowd counting on highly congested scenes.
\newblock {\em arXiv preprint arXiv:1612.00220}, 2016.

\bibitem{pan2017spatial}
X.~Pan, J.~Shi, P.~Luo, X.~Wang, and X.~Tang.
\newblock Spatial as deep: Spatial cnn for traffic scene understanding.
\newblock {\em arXiv preprint arXiv:1712.06080}, 2017.

\bibitem{ranjan2018iterative}
V.~Ranjan, H.~Le, and M.~Hoai.
\newblock Iterative crowd counting.
\newblock {\em arXiv preprint arXiv:1807.09959}, 2018.

\bibitem{richter2017playing}
S.~R. Richter, Z.~Hayder, and V.~Koltun.
\newblock Playing for benchmarks.
\newblock In {\em Proceedings of the International conference on computer
  vision}, volume~2, 2017.

\bibitem{Richter_2016_ECCV}
S.~R. Richter, V.~Vineet, S.~Roth, and V.~Koltun.
\newblock Playing for data: {G}round truth from computer games.
\newblock In {\em Proceedings of the European Conference on Computer Vision},
  pages 102--118, 2016.

\bibitem{DBLP:conf/iccv/RodriguezSLA11}
M.~Rodriguez, J.~Sivic, I.~Laptev, and J.~Audibert.
\newblock Data-driven crowd analysis in videos.
\newblock In {\em {IEEE} International Conference on Computer Vision}, pages
  1235--1242, 2011.

\bibitem{ros2016synthia}
G.~Ros, L.~Sellart, J.~Materzynska, D.~Vazquez, and A.~M. Lopez.
\newblock The synthia dataset: A large collection of synthetic images for
  semantic segmentation of urban scenes.
\newblock In {\em Proceedings of the IEEE conference on computer vision and
  pattern recognition}, pages 3234--3243, 2016.

\bibitem{sam2017switching}
D.~B. Sam, S.~Surya, and R.~V. Babu.
\newblock Switching convolutional neural network for crowd counting.
\newblock In {\em Proceedings of the IEEE Conference on Computer Vision and
  Pattern Recognition}, volume~1, page~6, 2017.

\bibitem{shen2018crowd}
Z.~Shen, Y.~Xu, B.~Ni, M.~Wang, J.~Hu, and X.~Yang.
\newblock Crowd counting via adversarial cross-scale consistency pursuit.
\newblock In {\em Proceedings of the IEEE Conference on Computer Vision and
  Pattern Recognition}, pages 5245--5254, 2018.

\bibitem{shi2018crowd}
Z.~Shi, L.~Zhang, Y.~Liu, X.~Cao, Y.~Ye, M.-M. Cheng, and G.~Zheng.
\newblock Crowd counting with deep negative correlation learning.
\newblock In {\em Proceedings of the IEEE Conference on Computer Vision and
  Pattern Recognition}, pages 5382--5390, 2018.

\bibitem{simonyan2014very}
K.~Simonyan and A.~Zisserman.
\newblock Very deep convolutional networks for large-scale image recognition.
\newblock {\em arXiv preprint arXiv:1409.1556}, 2014.

\bibitem{sindagi2017cnn}
V.~A. Sindagi and V.~M. Patel.
\newblock Cnn-based cascaded multi-task learning of high-level prior and
  density estimation for crowd counting.
\newblock In {\em Proceedings of the IEEE International Conference on Advanced
  Video and Signal Based Surveillance}, pages 1--6, 2017.

\bibitem{sindagi2017generating}
V.~A. Sindagi and V.~M. Patel.
\newblock Generating high-quality crowd density maps using contextual pyramid
  cnns.
\newblock In {\em Proceedings of the IEEE International Conference on Computer
  Vision}, pages 1879--1888, 2017.

\bibitem{wang2018detecting}
Q.~Wang, M.~Chen, F.~Nie, and X.~Li.
\newblock Detecting coherent groups in crowd scenes by multiview clustering.
\newblock {\em IEEE Transactions on Pattern Analysis and Machine Intelligence},
  2018.

\bibitem{Qi2017Deep}
Q.~Wang, J.~Wan, and Y.~Yuan.
\newblock Deep metric learning for crowdedness regression.
\newblock {\em IEEE Transactions on Circuits and Systems for Video Technology},
  28(10):2633--2643, 2018.

\bibitem{wang2004image}
Z.~Wang, A.~C. Bovik, H.~R. Sheikh, and E.~P. Simoncelli.
\newblock Image quality assessment: from error visibility to structural
  similarity.
\newblock {\em IEEE transactions on image processing}, 13(4):600--612, 2004.

\bibitem{zeng2017multi}
L.~Zeng, X.~Xu, B.~Cai, S.~Qiu, and T.~Zhang.
\newblock Multi-scale convolutional neural networks for crowd counting.
\newblock In {\em Proceedings of the IEEE International Conference on Image
  Processing}, pages 465--469, 2017.

\bibitem{zhang2016data}
C.~Zhang, K.~Kang, H.~Li, X.~Wang, R.~Xie, and X.~Yang.
\newblock Data-driven crowd understanding: a baseline for a large-scale crowd
  dataset.
\newblock {\em IEEE Transactions on Multimedia}, 18(6):1048--1061, 2016.

\bibitem{zhang2015cross}
C.~Zhang, H.~Li, X.~Wang, and X.~Yang.
\newblock Cross-scene crowd counting via deep convolutional neural networks.
\newblock In {\em Proceedings of the IEEE Conference on Computer Vision and
  Pattern Recognition}, pages 833--841, 2015.

\bibitem{zhang2016single}
Y.~Zhang, D.~Zhou, S.~Chen, S.~Gao, and Y.~Ma.
\newblock Single-image crowd counting via multi-column convolutional neural
  network.
\newblock In {\em Proceedings of the IEEE conference on Computer Vision and
  Pattern Recognition}, pages 589--597, 2016.

\bibitem{zhu2017unpaired}
J.-Y. Zhu, T.~Park, P.~Isola, and A.~A. Efros.
\newblock Unpaired image-to-image translation using cycle-consistent
  adversarial networks.
\newblock {\em arXiv preprint}, 2017.

\end{thebibliography}
	}

	\newpage
	\qquad
	\newpage
	
	\onecolumn
	
	\Large{\textbf{Supplementary}}
	\\
	\normalsize

	This file provides some additional information from three perspective: dataset, supervised and domain adaptation methods, which correspond to the Section 3, 4 and 5 in the paper. 
	\section{GCC Dataset}
	\subsection{Exemplars of GCC Dataset}
	
\begin{figure*}[h]
		\centering
		\includegraphics[width=0.95\textwidth]{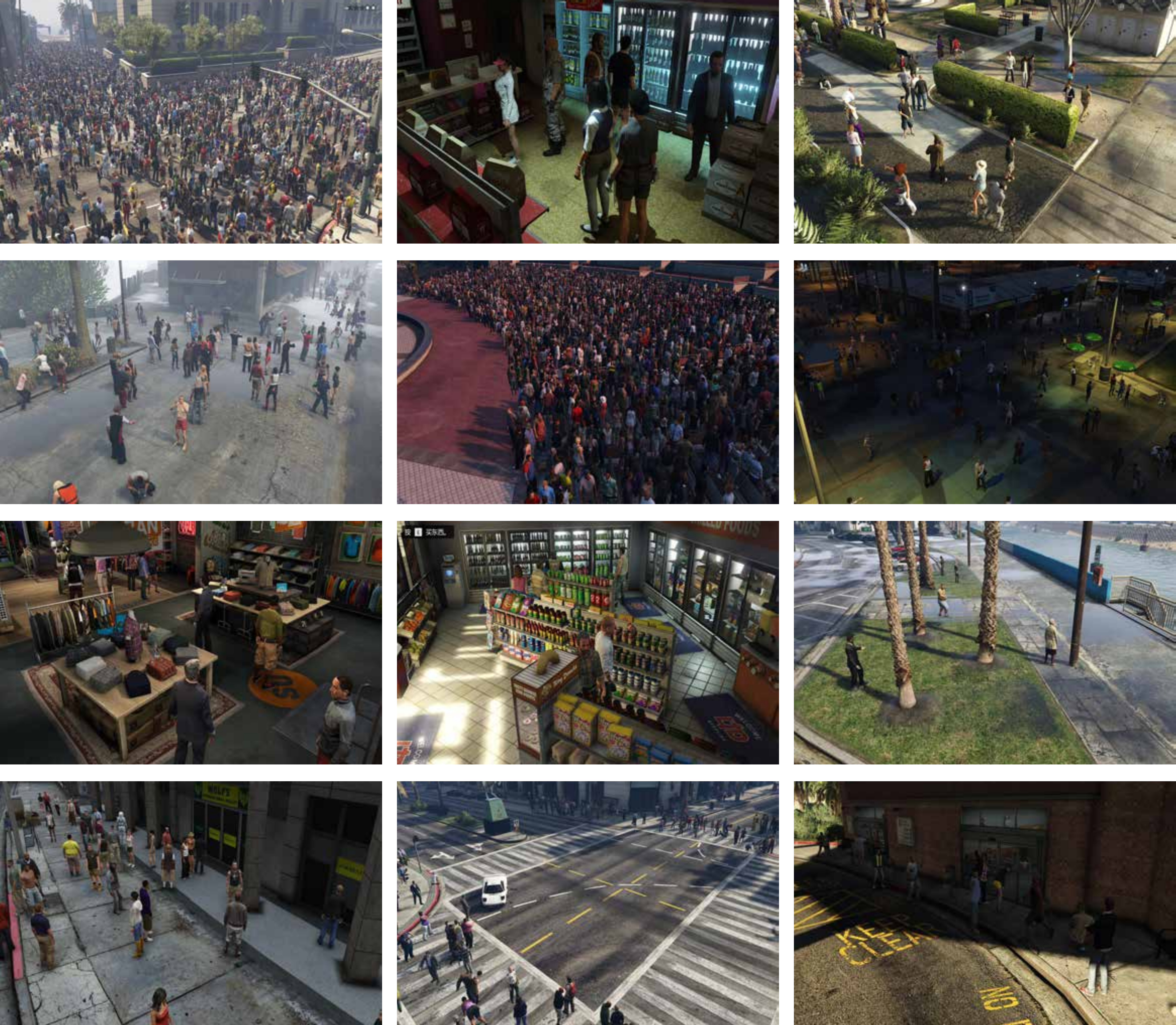}
		\caption{The exemplars of synthetic crowd scenes from the proposed GCC dataset. }\label{Fig-show}
	\end{figure*}

	For a deeper understanding GCC dataset, some typical crowd scenes are shown in Fig. \ref{Fig-show}.
	
	\subsection{Information Provided by GCC}
	
	For each scene, the complete camera parameters in the virtual world are provided: position coordinates, height, pitch/yaw angle and field of view. In addition, we also provide the Region of Interest (ROI) for placing person models, which is represented by a polygon region. According to the area of ROI, we assign a capacity label from  9 levels for each scene. Based on aforementioned parameters, all scenes in GCC dataset can be easily reproduced.
	
	For one specific crowd image, in addition to coordinates of head locations, we also provide its capturing time in 24h, weather condition and binary crowd segmentation map.
	
	\subsection{100 Locations in GTA5 World}
	
\begin{figure*}[!htbp]
		\centering
		\includegraphics[width=0.75\textwidth]{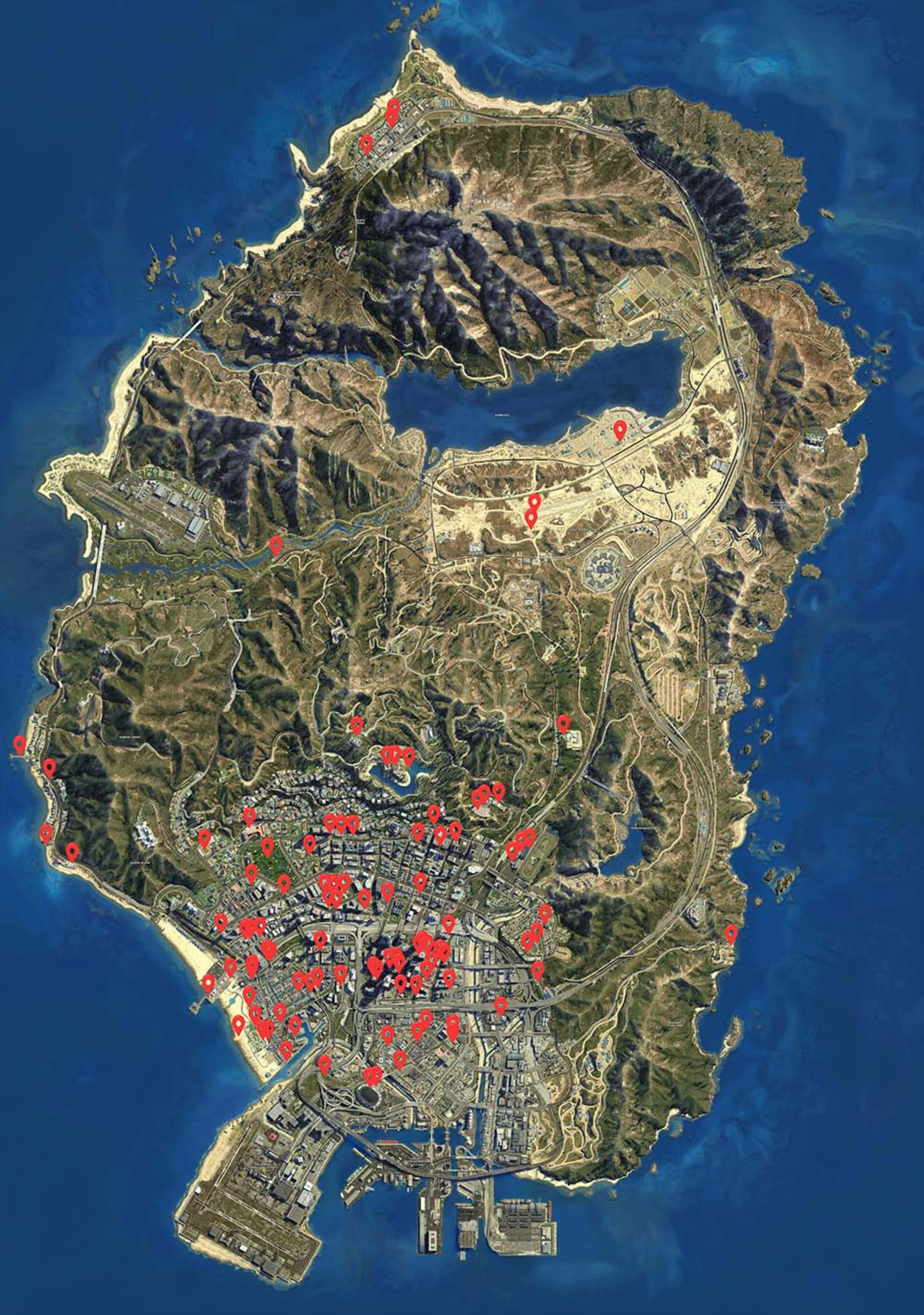}
		\caption{The demonstration of selected 100 locations in GTA5 world. }\label{Fig-loc}
	\end{figure*}

	Fig. \ref{Fig-loc} demonstrates the position of each location in GTA5 world. In general, our locations are mainly concentrated in the urban area.
	
	\newpage
	\section{Supervised Crowd Counting}
	
	\subsection{Configuration Details of the Proposed Networks in this Paper}
	
	Table \ref{Table-net} explains the configurations of FCN, SFCN and SFCN\dag. In the table, ``k(3,3)-c256-s1-d2'' represents the convolutional operation with kernel size of $3 \times 3$, $256$ output channels, stride size of $1$ and dilation rate of $2$. Note that we modify the stride size to 1 in conv4\_x of ResNet-101 backbone, which makes conv4\_x output the feature maps with 1/8 size of the input image. Other architecture settings fully follow the original VGG-16 and ResNet-101.
	
	\begin{table*}[htbp]
		
		\centering
		\small
		\caption{The network architectures of FCN, SFCN and SFCN\dag.}
		\begin{tabular}{|p{3cm}<{\centering}|p{3cm}<{\centering}|p{6cm}<{\centering}|}
			\whline
			FCN	&SFCN&SFCN\dag \\
			\whline
			\multicolumn{2}{|c|}{\textbf{VGG-16 backbone}} & \textbf{ResNet-101 backbone}  \\
			\multicolumn{2}{|c|}{conv1: [k(3,3)-c64-s1] $\times$ 2} & conv1: k(7,7)-c64-s2 \\
			\multicolumn{2}{|c|}{...} &...\\
			\multicolumn{2}{|c|}{conv3: [k(3,3)-c512-s1] $\times$ 3} &conv4\_x: $\left[ \begin{array}{l}
			k(1,1) - c256 - s1\\
			k(3,3) - c256 - s1\\
			k(1,1) - c1024 - s1
			\end{array} \right] \times 23$ \\
			\hline
			- &\multicolumn{2}{c|}{\textbf{Dilation Convolution}} \\
			- & \multicolumn{2}{c|}{k(3,3)-c512-s1-d2}\\
			- & \multicolumn{2}{c|}{k(3,3)-c512-s1-d2}\\
			- & \multicolumn{2}{c|}{k(3,3)-c512-s1-d2}\\
			- & \multicolumn{2}{c|}{k(3,3)-c256-s1-d2}\\
			- & \multicolumn{2}{c|}{k(3,3)-c128-s1-d2}\\
			- & \multicolumn{2}{c|}{k(3,3)-c64-s1-d2}\\
			\hline
			- &\multicolumn{2}{c|}{\textbf{Spatial Encoder}}\\
			- &\multicolumn{2}{c|}{down: k(1,9)-c64-s1}\\
			- &\multicolumn{2}{c|}{up: k(1,9)-c64-s1}\\
			- &\multicolumn{2}{c|}{left-to-right: k(9,1)-c64-s1}\\
			- &\multicolumn{2}{c|}{right-to-left: k(9,1)-c64-s1}\\
			\hline
			\multicolumn{3}{|c|}{\textbf{Regression Layer}}   \\
			\multicolumn{3}{|c|}{k(1,1)-c1-s1}   \\
			\multicolumn{3}{|c|}{upsample layer: $\times$8}   \\
			\whline
		\end{tabular}
		\label{Table-net}
	\end{table*}
	
	
	\subsection{Performance of SFCN on GCC}
	
	Table \ref{Table-gcc} lists the results on GCC dataset. The models are evaluated using standard Mean Absolute Error (MAE) and Mean Squared Error. In the table, ``Average'' denotes the average value of each class. 
	
	\begin{table*}[htbp]
		
		\centering
		\caption{Results of SFCN on GCC dataset (MAE/MSE).}
		\small
		\setlength{\tabcolsep}{1.5mm}{
			\begin{tabular}{c|c|c|c|c|c|c|c|c|c|c}
				\multicolumn{11}{c}{\small{Performance of SFCN in each class}}\\
				\hline
				Method  & \textbf{Average} & 0$\sim$10 & 0$\sim$25  & 0$\sim$50 & 0$\sim$100 & 0$\sim$300 & 0$\sim$600 & 0$\sim$1k & 0$\sim$2k & 0$\sim$4k \\
				\hline
				random  &\textbf{28.7/46.2}  &6.5/8.8  &8.5/14.2  &6.8/10.2  &5.7/8.7  &11.5/16.1  &20.8/27.9  &32.9/46.9  &52.1/91.5  &113.8/191.8  \\
				\hline
				cross-camera    &\textbf{47.0/73.0}  &13.7/23.3  &14.7/18.3  &10.3/13.6  &11.1/14.0  &17.6/27.5  &21.8/29.1  &57.3/73.4  &96.2/165.0  &180.6/293.3  \\
				\hline
				cross-location    &\textbf{58.4/87.2}  &4.7/4.9  &7.8/13.5  &11.0/13.2  &11.4/13.3  &17.2/24.5  &20.9/28.3  &18.6/26.3  &138.3/232.3  &295.8/428.6 \\
				\hline
				
			\end{tabular}
		}
		\small
		\setlength{\tabcolsep}{1.3mm}{
			\begin{tabular}{c|c|c|c|c|c|c|c|c|c}	
				\multicolumn{10}{c}{\small{Performance of SFCN at different time periods}}\\
				\hline
				Method  & \textbf{Average} & 0$\sim$3 & 3$\sim$6  & 6$\sim$9 & 9$\sim$12 & 12$\sim$15 & 15$\sim$18 & 18$\sim$21 & 21$\sim$24  \\
				\hline
				random   &\textbf{41.4/96.7}  &54.5/110.4  &49.5/135.5  &29.1/72.2  &29.6/76.5  &33.4/64.2  &34.2/80.2  &47.2/87.7  &54.1/146.7  \\
				\hline
				cross-camera     &\textbf{63.7/147.4}  &77.9/192.0  &72.1/222.8  &52.6/113.9  &41.6/101.8  &70.7/144.0  &54.8/136.2  &78.5/147.9  &60.9/121.1   \\
				\hline
				cross-location    &\textbf{97.8/228.4}  &104.7/216.4  &138.8/308.2  &62.6/164.7  &81.3/209.8  &77.8/174.7  &94.7/235.9  &122.6/250.2  &100.1/267.2    \\
				\hline
				
			\end{tabular}
		}
		\small
		\setlength{\tabcolsep}{1.65mm}{
			\begin{tabular}{c|c|c|c|c|c|c|c|c}	
				\multicolumn{9}{c}{\small{Performance of SFCN under different weathers}}\\
				\hline
				Method  & \textbf{Average} & Clear & Clouds  & Rain & Foggy & Thunder & Overcast & Extra Sunny   \\
				\hline
				random   &\textbf{40.8/92.5}  &35.1/84.0  &36.0/64.8  &43.7/83.4  &58.2/167.6  &45.2/86.8  &34.2/84.9  &33.5/76.3  \\
				\hline
				cross-camera       &\textbf{68.3/155.6}  &54.4/130.9  &62.5/122.5  &87.8/208.3  &70.2/163.2  &73.4/163.1  &71.1/172.1  &58.5/129.1    \\
				\hline
				cross-location     &\textbf{106.8/246.2}  &76.1/185.2  &88.7/196.0  &128.2/286.8  &160.2/413.1  &117.7/232.8  &84.8/193.2  &92.1/216.4   \\
				\hline
				
			\end{tabular}
		}\label{Table-gcc}
	\end{table*}
	
	From the performance of the three aspects (random, cross-camera and cross-location splitting), both MAE and MSE are increased, which means the difficulty of three tasks is rising in turn. From the first table, the performance of small scenes is better than that of large scenes. The main reason is: the count ranges of the latter are far greater than that of the former, which causes that the former's errors become larger. The second table shows that the daytime scenes are easier to count the number of people than the night scenes. Similarly, from the third table, we also find the clear, cloud, overcast and extra sunny scenes are easier than the rain, foggy and thunder scenes.

	\section{Crowd Counting via Domain Adaptation}
	\subsection{Scene Regularization in Domain Adaptation}
	
	In the paper, we introduce Scene Regularization (SR) to select the proper images to avoid negative adaptation. This is not an elaborate selection but a coarse data filter. Here, Table \ref{Table-filter} shows the concrete filter condition for adaptation to the five real datasets.
	
	\begin{table*}[htbp]
		
		\centering
		\caption{Filter condition on five real datasets.}
		\begin{tabular}{|c|c|c|c|c|c|}
			\hline
			Target Dataset  & level & time & weather & count range & ratio range\\
			\hline
			SHT A  &4,5,6,7,8 & 6:00$\sim$19:59 & 0,1,3,5,6 &25$\sim$4000 & 0.5$\sim$1 \\
			\hline
			SHT B  &1,2,3,4,5 & 6:00$\sim$19:59 & 0,1,5,6 &10$\sim$600 & 0.3$\sim$1 \\
			\hline
			UCF\_CC\_50  &5,6,7,8 & 8:00$\sim$17:59 & 0,1,5,6 &400$\sim$4000 & 0.6$\sim$1\\
			\hline
			UCF-QNRF  &4,5,6,7,8 & 5:00$\sim$20:59 & 0,1,5,6 &400$\sim$4000 & 0.6$\sim$1\\
			\hline
			WorldExpo'10  &2,3,4,5,6 & 6:00$\sim$18:59 & 0,1,5,6 &0$\sim$1000 & 0$\sim$1\\
			\hline
		\end{tabular}
		\label{Table-filter}
	\end{table*}
	
	In Table \ref{Table-filter}, ratio range means that the numbers of people in selected images should be in a specific range. For example, during adaptation to SHT A, there is a candidate image with level 0$\sim$4000, containing 800 people. According to the ratio range of 0.5$\sim$1, since 800 is not in 2000$\sim$4000 (namely 0.5*4000 $\sim$1*4000), the image can not be selected. In other words, ratio range is a restriction in terms of congestion. 
	
	Other explanations of Arabic numerals in the table is listed as follows:
	
	\textbf{Level Categories} 0: 0$\sim$10, 1: 0$\sim$25, 2: 0$\sim$50, 3: 0$\sim$100, 4: 0$\sim$300, 5: 0$\sim$600, 6: 0$\sim$1k, 7: 0$\sim$2k and 8: 0$\sim$4k. 
	
	\textbf{Weather Categories} 0: clear, 1: clouds, 2: rain, 3: foggy, 4: thunder, 5: overcast and 6: extra sunny. 
	
	\subsection{Visualization Comparison of Cycle GAN and SE Cycle GAN}
	
\begin{figure*}[htbp]
		\centering
		\includegraphics[angle=270,width=0.5\textwidth]{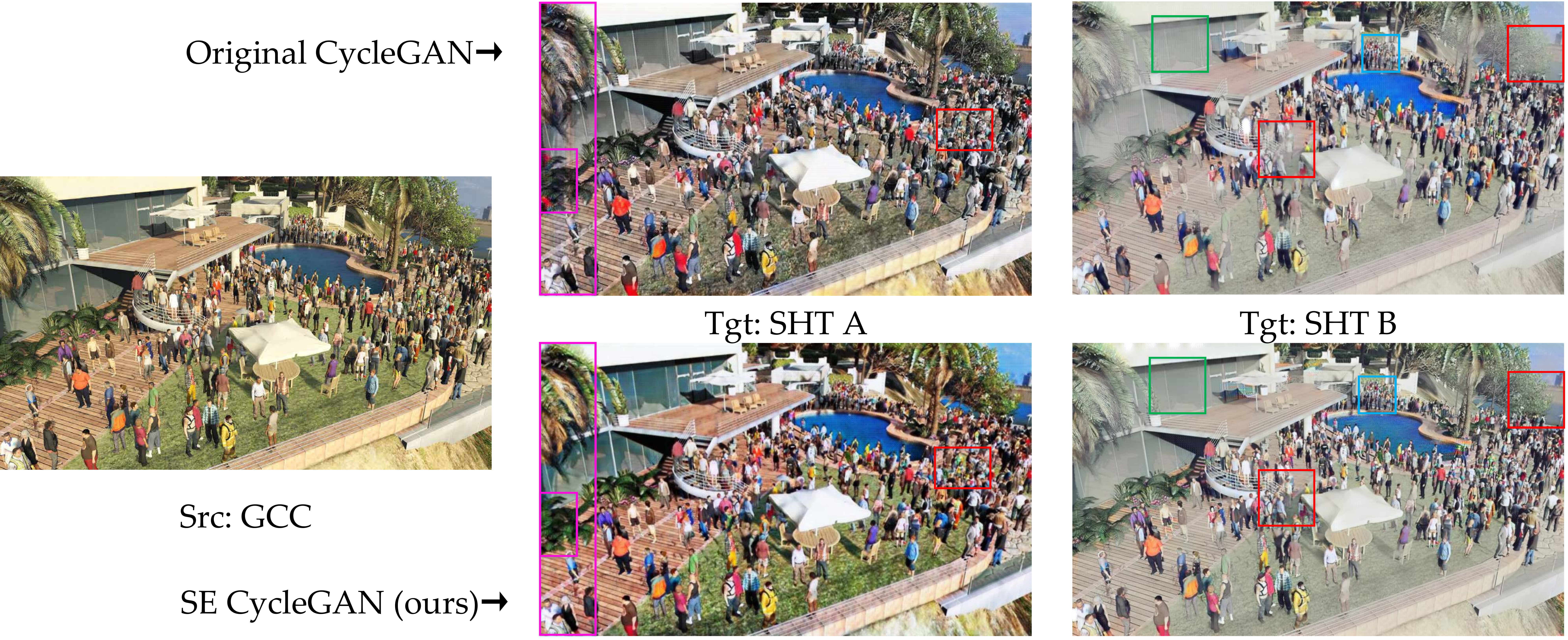}
		\caption{The exemplars of translated images. }\label{Fig-com-1}
	\end{figure*}

\begin{figure*}[htbp]
		\centering
		\includegraphics[angle=270,width=0.5\textwidth]{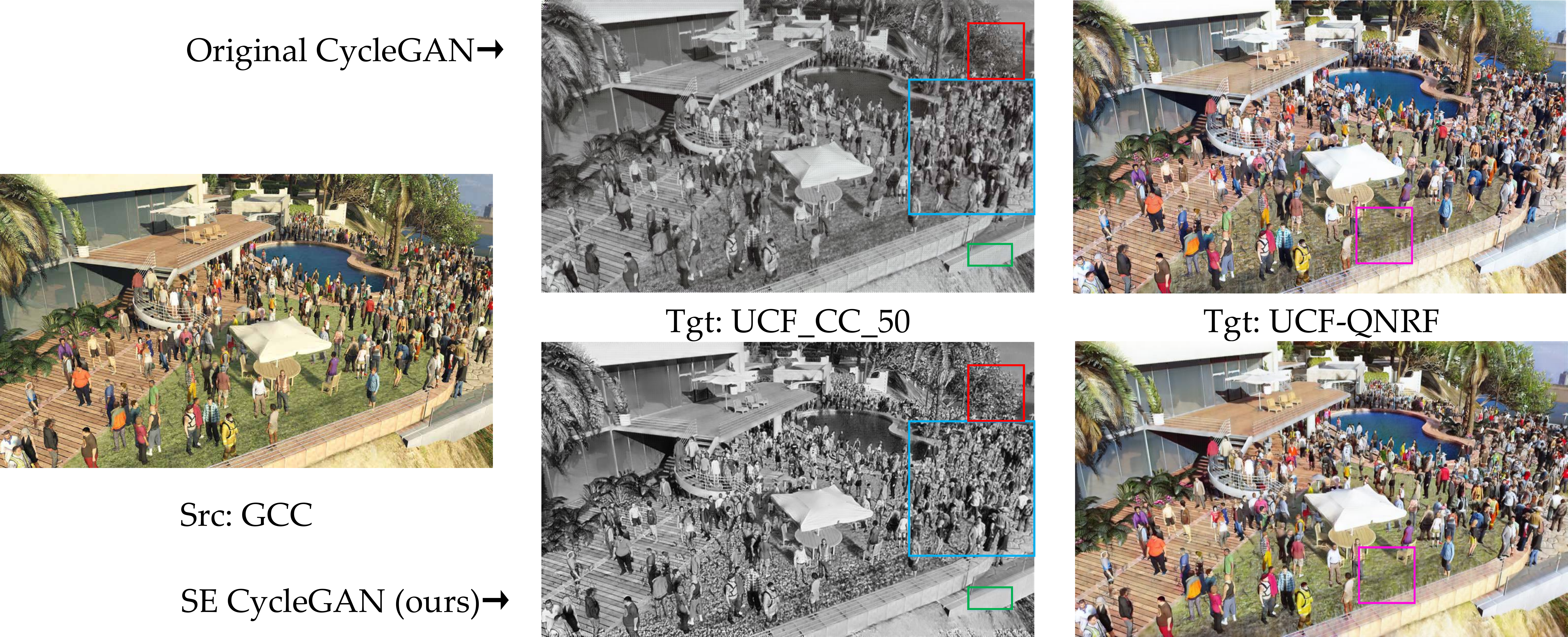}
		\caption{The exemplars of translated images. }\label{Fig-com-2}
	\end{figure*}

\begin{figure*}[htbp]
		\centering
		\includegraphics[angle=270,width=0.55\textwidth]{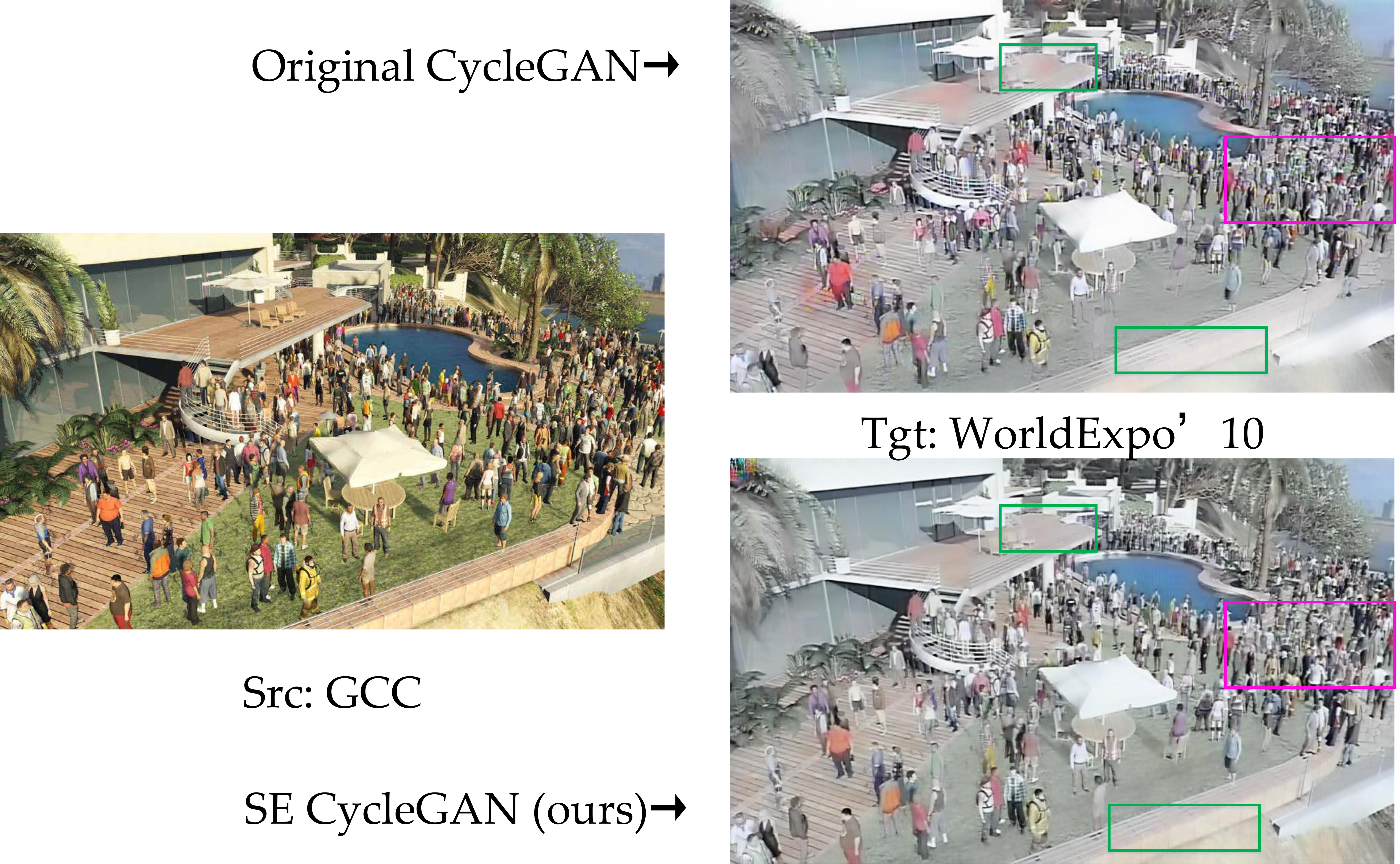}
		\caption{The exemplars of translated images. }\label{Fig-com-3}
	\end{figure*}

	Fig. \ref{Fig-com-1}, \ref{Fig-com-2} and \ref{Fig-com-3} demonstrate the translated images from GCC to the five real-world datasets. ``Src'' and ``Tgt'' represent the source domain (synthetic data) and target domain (real-world data). The top column shows the results of the original Cycle GAN and the bottom is the results of the proposed SE Cycle GAN.
	
	We compare some obvious differences between Cycle GAN and SE Cycle GAN (ours) and mark them up with rectangular boxes. To be specific, ours can produce more consistent image than the original Cycle GAN in the green boxes. As for the red boxes, Cycle GAN loses more texture features than ours. For the purple boxes, we find that Cycle GAN produces some abnormal color values, but SE Cycle GAN performs better than it. For the regions covered by blue boxes, SE Cycle GAN maintains the contrast of the original image than Cycle GAN in a even better fashion. 
	
	In general, from a visualization results, the proposed SE Cycle GAN generates more high-quality crowd scenes than the original Cycle GAN.

\end{document}